\documentclass[lettersize,journal]{IEEEtran}
\usepackage{amsmath,amsfonts}
\usepackage{algorithmic}
\usepackage{algorithm}
\usepackage{array}
\usepackage[font=normalsize,labelfont=normalsize,caption=false]{subfig}
\usepackage{textcomp}
\usepackage{stfloats}
\usepackage{url}
\usepackage{verbatim}
\usepackage{graphicx}
\usepackage{booktabs}
\usepackage{multirow}
\usepackage{cite}
\usepackage[table,xcdraw]{xcolor}
\usepackage{subfloat}
\usepackage{float}
\usepackage{makecell}
\usepackage[colorlinks=true,
citecolor=green!80!black,
linkcolor=red,
anchorcolor=black,
urlcolor=magenta]{hyperref}
\newcommand{\eat}[1]{}
\newcommand{\tb}\textbf
\newcommand{\myrule}{\specialrule{.1em}{.0ex}{.0ex}}
\usepackage[normalem]{ulem}
\newcommand{\rounda}[1]{\textcolor{black}{#1}}
\newcommand{\roundb}[1]{\textcolor{black}{#1}}

\useunder{\uline}{\ul}{}

\begin{document}

\title{MSFlow: Multi-Scale Flow-based Framework for Unsupervised Anomaly Detection}


	\author{Yixuan Zhou, Xing Xu, Jingkuan Song, Fumin Shen, Heng Tao Shen ~\IEEEmembership{Fellow,~IEEE}
	\thanks{This work was supported in part by National Key Research and Development Program of China (No. 2018AAA0102200). (Corresponding author: Heng Tao Shen)}
	\thanks{Y. Zhou, X. Xu, J. Song, F. Shen are with the Center for Future Media and School of Computer Science and Engineering, University of Electronic Science and Technology of China, Chengdu 611731, China (E-mail: yxzhou@std.uestc.edu.cn; xing.xu@uestc.edu.cn, fumin.shen@gmail.com, jingkuan.song@gmail.com).}
	\thanks{H. T. Shen is with the Center for Future Media and School of Computer Science and Engineering, University of Electronic Science and Technology of China, Chengdu 611731, China and also with Peng Cheng Laboratory, Shenzhen 518055, China (Email: shenhengtao@hotmail.com)}
}


\markboth{IEEE Transactions on Neural Networks and Learning Systems, Vol. XX, No. XX, 2023 (under review)}%
{Shell \MakeLowercase{\textit{et al.}}: A Sample Article Using IEEEtran.cls for IEEE Journals}

\maketitle

\begin{abstract}
  Unsupervised anomaly detection (UAD) attracts a lot of research interest and drives widespread applications, where only anomaly-free samples are available for training. Some UAD applications intend to locate the anomalous regions further even without any anomaly information. 
  Although the absence of anomalous samples and annotations deteriorates the UAD performance, an inconspicuous yet powerful statistics model, the normalizing flows, is appropriate for anomaly detection and localization in an unsupervised fashion. The flow-based probabilistic models, only trained on anomaly-free data, can efficiently distinguish unpredictable anomalies by assigning them much lower likelihoods than normal data.
  Nevertheless, the size variation of unpredictable anomalies introduces another inconvenience to the flow-based methods for high-precision anomaly detection and localization. To generalize the anomaly size variation, we propose a novel \textbf{M}ulti-\textbf{S}cale \textbf{Flow}-based framework dubbed \textit{MSFlow} composed of asymmetrical parallel flows followed by a fusion flow to exchange multi-scale perceptions. Moreover, different multi-scale aggregation strategies are adopted for image-wise anomaly detection and pixel-wise anomaly localization according to the discrepancy between them. The proposed MSFlow is evaluated on three anomaly detection datasets, significantly outperforming existing methods. Notably, on the challenging MVTec AD benchmark, our MSFlow achieves a new state-of-the-art with a detection AUORC score up to 99.7\%, localization AUCROC score of 98.8\% and PRO score of 97.1\%. The reproducible code is available at \url{https://github.com/cool-xuan/msflow}.
\end{abstract}


\section{Introduction}
Anomaly detection (AD) is a crucial and challenging field of research with broad applications such as fraud detection \cite{dal2017credit}, violence detection \cite{sultani2018real}, medical diagnostics \cite{paluru2021anam} and industrial defect detection \cite{bergmann2019mvtec}. 
Furthermore, many applications such as industrial defect detection are not only satisfied with just discriminating anomalies but also intend to locate the anomaly regions. Guided by anomaly localization, the manufacturers can optimize the industrial production process to improve the qualification rate and save costs. 
However, anomalies are rare, unpredictable, and diverse in real-world scenarios. It is impractical to collect a dataset containing all anomalous types. 
Annotating anomalous data is also costly, especially for pixel-wise localization annotations. Therefore, it is unavailable to collect labeled data to support the full-supervision training for anomaly detection. 
To address this inconvenience, 
unsupervised anomaly detection (UAD) collects easily accessible anomaly-free data for training, and anomalies are discriminated if there is any deviation from anomaly-free data. 

\begin{figure}
  \centering
  \includegraphics[width=\linewidth]{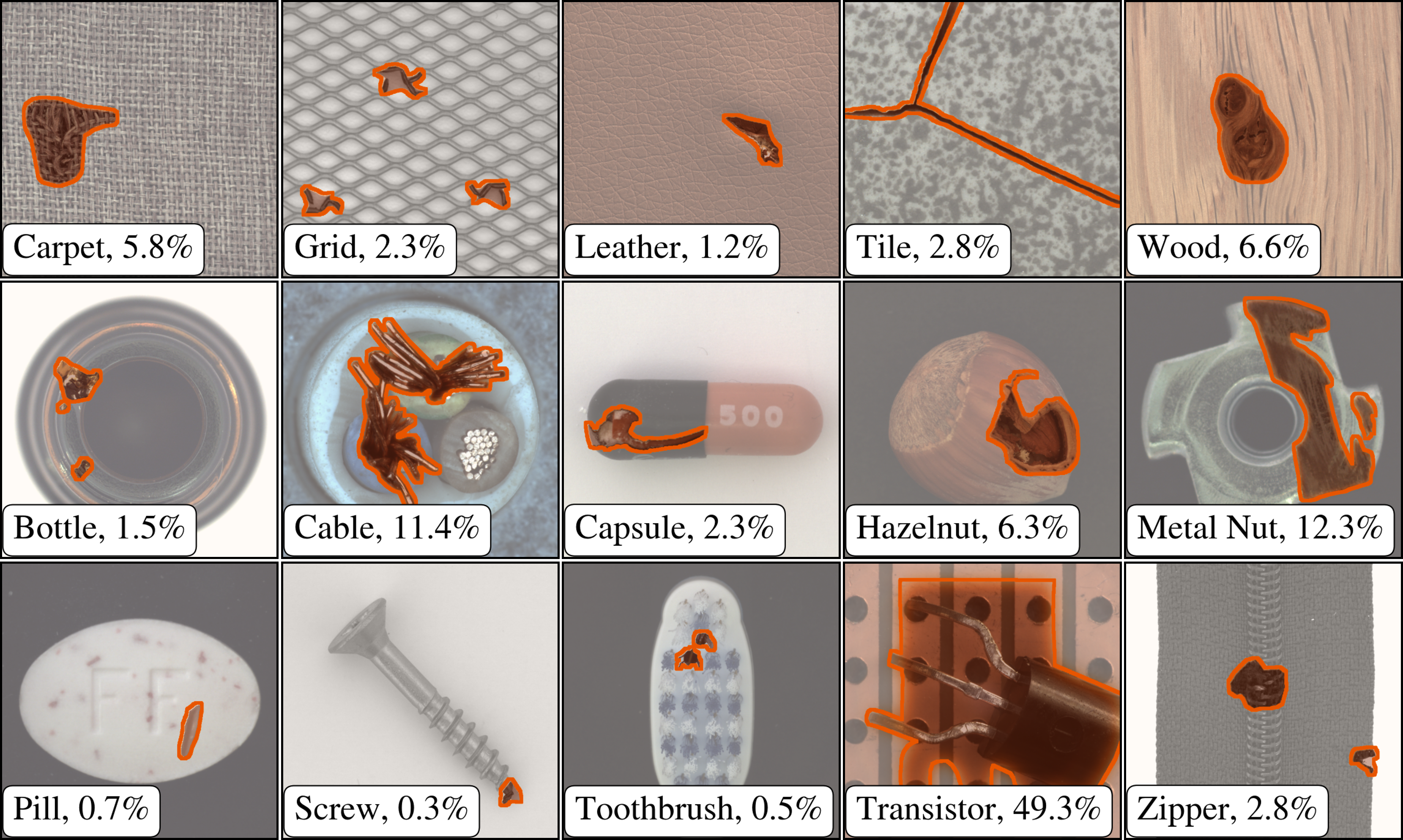} 
  \caption{Defect examples of all classes in MVTec AD benchmark. The percentages represent the area proportion of the defective regions.}
  \label{fig:defect-examples}
\end{figure}

The unsupervised fashion eases the data collection while intensifying the complication of anomaly localization. 
Most existing methods address this challenge via computing the pixel-wise reconstruction errors \cite{bergmann2019mvtec, zavrtanik2021riad, marimont2021anomaly, ristea2021self, deng2022distillation} and clustering pixel-wise or region-wise anomaly-free samples to distinguish the anomalous regions \cite{defard2021padim, reiss2021panda, cohen2020spade, roth2021patchcore}. Recently, \cite{rudolph2021differnet} first proposed to apply the normalizing flows \cite{rezende2015variational} to estimate the likelihood of the entire image for the image-wise anomaly detection. CFlow-AD \cite{gudovskiy2022cflow} extended the normalizing flows to the pixel-wise anomaly localization by estimating the likelihood for each feature vector positioned in the feature map. These flow-based methods reveal impressive effectiveness in unsupervised anomaly detection and localization. 

However, the unpredictable size variation of diverse anomalies remains an obstacle for flow-based methods to achieve high-precision detection and localization. 
As illustrated in Fig. \ref{fig:defect-examples}, the defect size varies enormously across industrial classes in the MVTec AD benchmark specialized for industrial defect detection and localization. 
The defect region of the \textit{misplaced transistor} covers nearly half (49.3\%) of the entire image, while the one of \textit{pill}, \textit{screw} or \textit{toothbrush} only covers a few pixels. For this issue, CSFlow \cite{rudolph2022csflow} proposed a cross-scale flow module to further exchange information captured on different scales, which is beneficial for the adaptation to anomalies with various sizes. However, the CSFlow only targets anomaly detection without considering localization, and its simple architecture is burdensome and inefficient in information exchange.

To tackle the size variation, we propose a novel \textbf{M}ulti-\textbf{S}cale \textbf{Flow}-based framework dubbed \textbf{MSFlow} tailored for unsupervised anomaly detection, as shown in Fig. \ref{fig:framework}. We perform multi-scale optimization of the flow-based methods in the following three aspects: 

1) \textit{Multi-scale Feature Pyramid Extraction}. Different from \cite{rudolph2021differnet,rudolph2022csflow} using the featurized image pyramid or \cite{gudovskiy2022cflow} extracting high-level but low-resolution feature maps, we build our multi-scale feature pyramid with the activation outputs of the low-level stages.
The pixel-wise anomaly localization focuses on spatial structure consistency, and global anomaly detection is highly dependent on the perception of local anomalous regions.
Therefore, the low-level feature pyramid with a larger scale and more spatial details are more favorable for this specific task. 

2) \textit{Multi-scale Flow Model.} Our multi-scale flow model not only independently transforms the feature map of each scale by respective \textit{parallel flows} but also employs \textit{a fusion flow} to fuse all feature maps for multi-scale perceptions exchange. In particular, our multi-scale parallel flows are stacked in an asymmetrical architecture, where the parallel flows applied on the higher-level feature map comprise more flow blocks. Our lightweight fusion flow shares a similar mechanism with CSFlow\cite{rudolph2022csflow} but is much more efficient in information exchange.

3) \textit{Multi-scale Outputs Aggregation.} We propose different multi-scale aggregation strategies for image-wise anomaly detection and pixel-wise anomaly localization considering the inherent discrepancy between these two subtasks.
Specifically, the addition aggregation maintaining the multi-scale properties is adopted to calculate the pixel-wise anomaly score map, while the multi-scale likelihood maps are aggregated by multiplication to suppress the noise before the image-wise anomaly score calculation. Furthermore, the mean of the largest $K$ scores in the multiplication-aggregated anomaly score map is treated as the global anomaly score.
Compared with the maximum \cite{zavrtanik2021riad,gudovskiy2022cflow,roth2021patchcore} or the mean \cite{rudolph2021differnet,rudolph2022csflow} widely used in existing methods, the mean of the top$K$ pixel-wise anomaly scores is more robust and sensitive to small defects.

To showcase the superiority of our method, the proposed MSFlow is compared with existing methods
on the challenging MVTec AD benchmark \cite{bergmann2019mvtec}.
Our MSFlow achieves state-of-the-art (SOTA) no matter either image-wise anomaly detection (AUCROC 99.7\%) or pixel-wise anomaly localization (AUCROC 98.8\% and PRO 97.1\%) on this industrial defect detection benchmark. 
The proposed MSFlow is also applied to another image anomaly detection dataset Magnetic Tile Defects (MTD) dataset \cite{huang2020mtd} and a non-image violence detection dataset mini Shanghai Tech Campus (mSTC) \cite{luo2017stc} to verify its generality. Our MSFlow achieves comparable accuracy on these two datasets as well. The comprehensive ablation studies highlight the effectiveness of the proposed components in the MSFlow.

Our main contributions can be summarized as follows:
\noindent{
\begin{itemize}
    \setlength{\itemsep}{0pt}
    \setlength{\parsep}{0pt}
    \setlength{\parskip}{3pt}
    \item To generalize the variation of anomaly size, we explore the multi-scale property of the flow-based method for unsupervised anomaly detection.
    A \textbf{M}ulti-\textbf{S}cale \textbf{Flow}-based framework (\tb{MSFlow}) is presented as our implementation.
    \item The asymmetrical parallel flows architecture in our MSFlow achieves the trade-off of performance and efficiency. A lightweight fusion flow further exchanges information of different scales and receptive fields to boost the flow model's generalizability of the anomaly size variation.
    \item Different multi-scale aggregation strategies are employed for image-wise anomaly detection and pixel-wise anomaly localization. Moreover, the mean of the top$K$ anomaly scores in the entire pixel-wise anomaly map is treated as the final image-wise anomaly score, which is robust and sensitive to small anomalous regions.
\end{itemize}}

The rest part of the paper is organized as follows. Section \ref{sec:related-work} introduces the previous works related to ours, followed by a brief introduction of the normalizing flows in Section \ref{sec:theory-background}. Section \ref{sec:method} provides implementation details about our MSFlow. Extensive experiments and ablation studies are provided in Section \ref{sec:experiment}, which exhibit the superiority of the proposed MSFlow. We conclude the paper in Section \ref{sec:conclusion}.


%

\section{Related Work} \label{sec:related-work}

\subsection{Unsupervised Anomaly Detection}
In real/practical scenarios, since only normal data are available, \textit{unsupervised anomaly detection} (UAD) is also named \textit{out-of-distribution} or \textit{one-class-calssification}. In an unsupervised fashion, normal samples and anomalies are discriminated through the inductive bias for normal training data. Predominant UAD approaches can be broadly summarized into three categories: \textit{reconstruction-based methods}, \textit{clustering-based methods}, and \textit{flow-based methods}, to which our method belongs.

\textit{Reconstruction-based methods} are the most widely used methodology not only in image anomaly detection but also in video violation detection \cite{sabokrou2017deepcascade, ergen2019unsupervised, luo2017stc}. These methods encode and reconstruct the normal data via generative models such as AutoEncoders (AEs) \cite{zhou2017anoae, akcay2018ganomaly,liu2020vevqe, zhou2021memorizing} and generative adversarial networks (GANs) \cite{perera2019ocgan}. The anomalies are distinguished based on the high reconstruction error assuming that the generative models trained on normal data perform poorly on the anomalies. However, such an assumption is invalid because the models trained with the strict MSE loss focus on pixel reconstruction and generalize to anomalies. To address this problem, there are some extensions based on weak restrictions like SSIM loss \cite{bergmann2018improving} or cosine similarity loss \cite{deng2022distillation}, generality degradation by memory modules \cite{gong2019memorizing} or codebook \cite{marimont2021anomaly}, and self-supervision such as rotation prediction \cite{hendrycks2019using}, transformed image restoration \cite{fei2020attribute} and mask inpainting \cite{zavrtanik2021riad}. Recently, some works \cite{li2021cutpaste, zavrtanik2021draem} synthesize pseudo defects on the anomaly-free data and train the reconstruction networks in a supervised way to boost the performance of anomaly localization. However, there is still no guarantee that the trained reconstructors only reconstruct the normal regions but not the defective regions. Besides, the reconstruction-based method can only detect the damage defects like cracks, scratches, or dents but not structural deformations like the \textit{misplaced transistor} in \ref{fig:defect-examples}.

\textit{Clustering-based methods} \cite{liu2021anomaly} build a reference gallery of normal data representations and then detect the anomalies based on the similarity with the reference gallery. \cite{ruff2018deepsvdd} use a deep neural network to learn the discriminative representations of normal data by SVM. However, most recent works utilize the feature extractors \cite{he2016resnet,zagoruyko2016wide,xie2017resnext} pre-trained on the large-scale external datasets \cite{deng2009imagenet} to estimate the representative features. Based on such extracted representations, $k$ nearest neighbors clustering \cite{bergman2020deepknn,reiss2021panda}, memory banks \cite{cohen2020spade} and bag-of-features approach \cite{defard2021padim} are used to build the reference gallery on normal data. The previous SOTA on the MVTec AD benchmark PatchCore \cite{roth2021patchcore} employs the kNN algorithm on patch-wise features to boost inference efficiency and constrain generalization to anomalies before building the memory bank. Nonetheless, the inference speed of clustering-based methods is still slow due to the time-consuming kNN algorithm as post-processing. Similar to reconstruct-based methods, the methods of this category also perform inferiorly on structural deformations.

\begin{figure*}[t]
  \centering
  \includegraphics[width=0.95\textwidth]{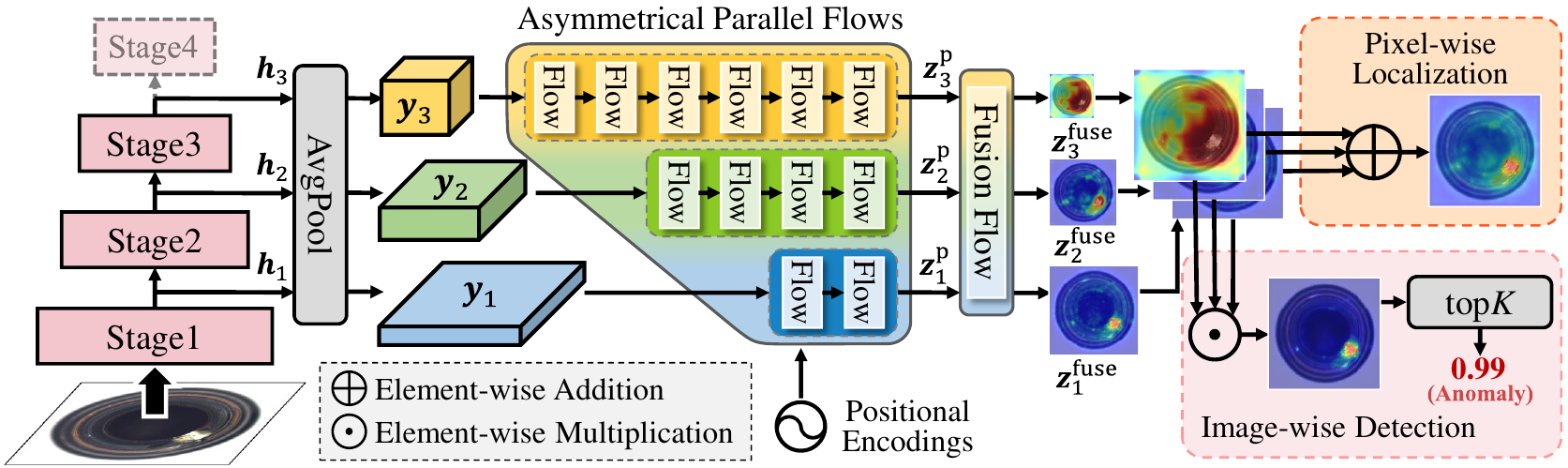}
  \caption{The framework of the MSFlow. The feature maps of the first 3 stages are extracted and $2\times$ downsampled by average pooling. The networks of stage4 are abandoned in our MSFlow, which is specifically drawn as a dashed box with a lighter color. The flow model in the MSFlow is composed of asymmetrical parallel flows and a fusion flow for information exchange. Finally, multiplication and addition aggregation are adopted for image-wise anomaly detection and pixel-wise anomaly localization, respectively.}
  \label{fig:framework}
\end{figure*}

\subsection{Normalizing Flows}
Compared with the well-known generative models VAEs \cite{kingma2013vaes} and GANs \cite{goodfellow2014gans}, the normalizing flows are inconspicuous but powerful. With the invertible property, the normalizing flows can transform any complex distribution into a tractable base distribution like the Gaussian distribution. To satisfy the invertibility of flow blocks, diverse implementations such as the autoregressive flow \cite{germain2015made} and the reverse autoregressive flow \cite{kingma2016improved} are proposed. Nevertheless, these two implementations are only effective in a single direction but costly in the other direction. The widespread Real-NVP \cite{dinh2016realnvp} implements the normalizing flows based on the affine coupling layers, which are efficient in both directions. Following the Real-NVP, \cite{kingma2018glow} further simplifies the flow architecture by introducing the invertible $1\times1$ convolutions to generalize any channel-wise permutation. They construct a large-scale Glow with numerous flow blocks for high-quality human face generation, achieving comparable results to GANs.

In addition to probabilistic modeling like image generation, the normalizing flows perform well in exact density estimation \cite{germain2015made,dinh2016realnvp} with invertibility. The training objective of normalizing flows for density estimation requires no annotation but only one pre-defined base distribution, commonly set as the Gaussian distribution. Therefore, the normalizing flows can be easily extended to UAD with their capability of explicitly estimating the probabilistic density of the normal data.

\subsection{Flow-based Methods for UAD} 
Flow-based methods have recently arisen in UAD, and our method belongs to this emerged methodology.
The methods of this category utilize the powerful normalizing flows to estimate the normal data's density, and the unseen anomalies are assigned with low likelihoods \cite{serra2019input}. 
However, \cite{kirichenko2020normalizing} reveals that the flow models trained on raw RGB images often assign even higher likelihoods to anomalies than the normal data. This puzzling result can be settled by applying the flows to the high-dimensional features instead of the raw image pixels.
With this revelation, DifferNet \cite{rudolph2021differnet} implements normalizing flows for the image-wise anomaly detection based on the extracted features.
To handle the defect size variation, DifferNet rescales the images to 3 scales and extracts the multi-scale feature maps through the same extractor. CFlow-AD \cite{rudolph2021differnet} builds its multi-scale feature pyramid with feature maps of different levels for efficiency and various receptive fields. Besides, CFlow-AD extends the normalizing flows to the pixel-wise anomaly localization by estimating the likelihood for the feature vector of each position.

However, CFlow-AD separates each position's feature vector and estimates its likelihood independently, neglecting spatial contextual information. The lack of context-awareness harms the global perception and suffers disconnected localization results. To remedy that, our MSFlow estimates the features in all positions parallelly and builds the flows with $3\times3$ convolutions to automatically learn the contextual information.
\rounda{
The detection-only method CSFlow \cite{rudolph2022csflow} develops a fully-convolutional cross-scale flow module to jointly process multi-scale feature maps. To take full advantage of the multi-scale perceptions, we build a fusion flow module inspired by the lightweight module\cite{yu2021lite} to exchange information across different scales. Our fusion flow plays the same role as the cross-scale flow module in CSFlow while is much more efficient in information exchange. }

\begin{figure}[t]
  \includegraphics[width=\linewidth]{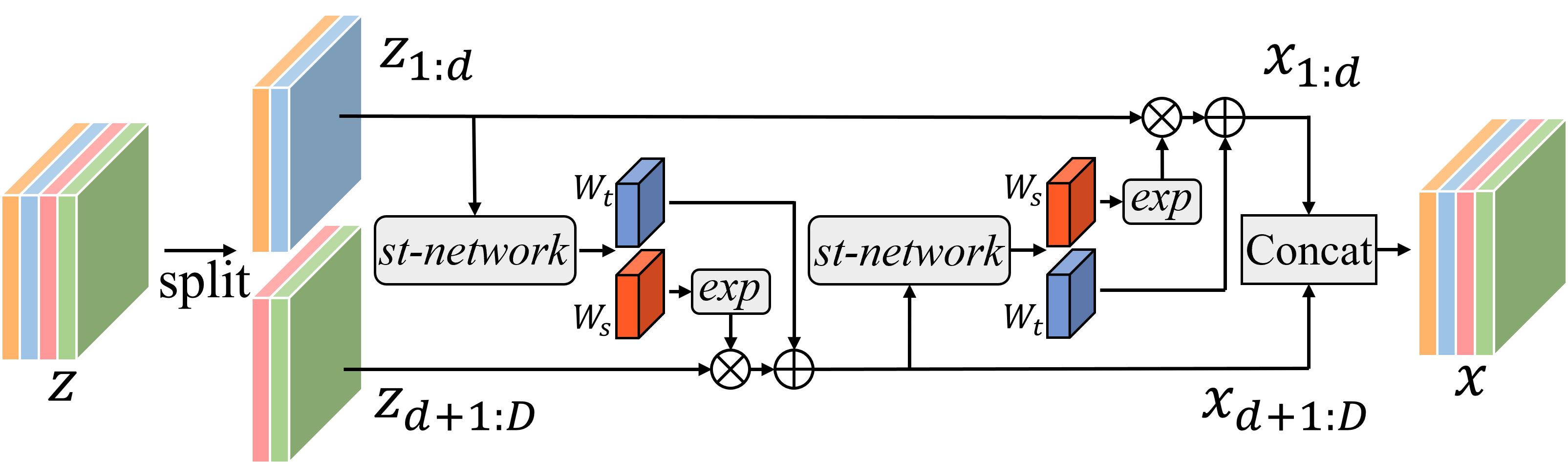}
  \caption{The pipeline of the flow block including two affine coupling layers. Each affine coupling layer encodes the halved features, and the encoded features are concatenated together as the output.}
  \label{fig:fusion-block}
\end{figure}

\section{Preliminary} \label{sec:theory-background}
Before presenting our method, we first introduce the theoretical background of the normalizing flows and the most common implementation: the normalizing flows based on affine coupling layers.

Unlike the widely-used generative models including GANs \cite{goodfellow2014generative, esser2021taming} and VAEs \cite{kingma2013vaes, razavi2019generating}, which learn the data distribution by a proxy adversarial task or maximizing the ELBO, the flow models explicitly estimate arbitrary data distribution through the bijective invertible normalizing flows. 
The flow model $f$ can transform arbitrary complex distribution $p_X(\boldsymbol{x})$ to a tractable base distribution $p_Z(\boldsymbol{z})$.
Thus, a bijection is built between the data sample $\boldsymbol{x}$ and the $\boldsymbol{z}$ through the flow model $f$ comprising a chain of $K$ flow blocks $\{f_1, f_2, ..., f_K\}$:
\begin{equation}
  \begin{aligned}
    \boldsymbol{x}=f(\boldsymbol{z})=& f_K \circ f_{K-1} \circ ... \circ f_1(\boldsymbol{z}); \\
    \boldsymbol{z}=f^{-1}(\boldsymbol{x})=& {f_1}^{-1} \circ {f_2}^{-1} \circ ... \circ {f_K}^{-1}(\boldsymbol{x}),
  \end{aligned}
\end{equation}
where $\circ$ denotes the cascade of flow blocks.
With the invertible property of the normalizing flows, the likelihood of the input data $\boldsymbol{x}$ can be estimated by a change of variables \cite{halmos2013measure}:
\begin{equation}
  p_X(\boldsymbol{x})=p_Z(\boldsymbol{z})\left\lvert \textrm{det}J_{f^{-1}}(\boldsymbol{x}) \right\rvert,
  \, \textrm{where} \, J_{f^{-1}}(\boldsymbol{x})= \frac{\partial f^{-1}}{\partial \boldsymbol{x}}.
\end{equation}
$J_{f^{-1}}(\boldsymbol{x})$ is the Jacobian of the reversed normalizing flows $f^{-1}$, and $\left\lvert\textrm{det}(J)\right\rvert$ denotes the absolute determinant of the Jacobian matrix $J$.
\rounda{Therefore, the optimization target of maximizing the likelihood of the implicit distribution $p_X(\boldsymbol{x})$ can be equivalently transformed to minimizing the negative log-likelihood $- \textrm{log}p_Z(\boldsymbol{z})$ of the accessible distribution $p_Z(\boldsymbol{z})$ along with the Jacobian determinant as follows:
\begin{equation}
  \begin{aligned}
  \max p_X(\boldsymbol{x}) &\Leftrightarrow \min \, - \textrm{log} \,p_X(\boldsymbol{x}) \\
  &\Leftrightarrow \min - \textrm{log}\,p_Z(\boldsymbol{z}) - \textrm{log}\left\lvert \textrm{det}J_{f^{-1}}(\boldsymbol{x}) \right\rvert.
  \end{aligned}
\end{equation}
For simplicity, the base distribution p$_Z(\boldsymbol{z})$ is commonly selected as a multivariate Gaussian distribution. The corresponding optimization target is formulated as follows:
\begin{equation} \label{equation:loss-function}
  \max p_X(\boldsymbol{x}) \Leftrightarrow \min \frac{\left\lVert \boldsymbol{z} \right\rVert_2^2}{2} - \textrm{log}\left\lvert \textrm{det}J_{f^{-1}}(\boldsymbol{x}) \right\rvert.
\end{equation}
}

According to the above, the flow block $f_i$ in the normalizing flows $f$ should satisfy two basic properties: 1) it can be easily invertible, and 2) its Jacobian determinant is computationally efficient.
There are diverse implementations \cite{germain2015made, van2016pixel, van2016wavenet, papamakarios2017masked, kingma2016improved} of normalizing flows, where the normalizing flows based on affine coupling layers are flexible and most widely used \cite{dinh2014nice, dinh2016realnvp,kingma2018glow,kirichenko2020normalizing}. 
In the normalizing flows of this type, as illustrated in Fig. \ref{fig:fusion-block}, the bijective flow block $f_\textrm{aff}$ is conveniently implemented by the affine coupling layer as follows:
\begin{equation}
  \begin{aligned}
  \textrm{forward:}&
  \begin{cases}
  \boldsymbol{x}_{1:d}  &= \boldsymbol{z}_{1:d}, \\
  \boldsymbol{x}_{d+1:D}&= \boldsymbol{z}_{d+1:D} \odot \textrm{exp}(s(\boldsymbol{z}_{1:d})) +t(\boldsymbol{z}_{1:d});
  \end{cases}
  \\
  \textrm{reverse:}&
  \begin{cases}
  \boldsymbol{z}_{1:d}  &= \boldsymbol{x}_{1:d},\\
  \boldsymbol{z}_{d+1:D}&= (\boldsymbol{x}_{d+1:D}-t(\boldsymbol{x}_{1:d}))\odot \textrm{exp}(-s(\boldsymbol{x}_{1:d})). \nonumber
  \end{cases}
  \end{aligned}
\end{equation}
$\boldsymbol{z}_{1:d}$ and $\boldsymbol{z}_{d+1:D}$ are halved parts of $z$ with channel dimensions $D$, $\boldsymbol{x}_{1:d}$ and $\boldsymbol{x}_{d+1:D}$ are two corresponding parts of $x$. The $s(\cdot )$ and $t(\cdot )$ functions are commonly implemented by the neural network (st-network for short), yielding the \textit{scale} weights $\boldsymbol{w}_s$ and \textit{shift} $\boldsymbol{w}_t$ weights. 

With the ingenious design of the affine coupling layer, its reversed Jacobian determinant can be easily calculated as follows:
\begin{equation}
  \textrm{log}\left\lvert \textrm{det}J_{{f_\textrm{aff}}^{-1}}(\boldsymbol{x}) \right\rvert = \sum s(\boldsymbol{x}_{1:d}).
\end{equation}
Our MSFlow also adopts such flexible normalizing flows and makes adaptive adjustments on the st-network according to the particularity of industrial defect detection with localization.

\section{Proposed Method} \label{sec:method}
As shown in Fig. \ref{fig:framework}, we first extract the multi-scale feature maps in the first $3$ stages as the input of the normalizing flows. 
The extracted multi-scale feature maps are then fed into their respective series of fully-convolutional parallel flows, followed by a fusion flow to exchange information of different scales and perceptive fields.
Finally, the multi-scale likelihood maps are aggregated by different aggregation strategies for the discrepancy between image-wise anomaly detection and pixel-wise anomaly localization. The mean of the top$K$ scores in the entire anomaly score map is treated as the image-wise anomaly score.

\rounda{
\subsection{Symbol Defintion}
We first introduce the formal definition of mathematical symbols used in the following statements. The bold lowercase letters $\boldsymbol{x}$, $\boldsymbol{y}$, $\boldsymbol{h}$, and $\boldsymbol{z}$ respectively refer to the input image, outputs of the extractor, the inputs and outputs of the flow model.
The normal lowercase letters $f$ and $g$ refer to the flow module and common network.
$st$ denotes the st-network in flow modules.
Particularly, the superscripts indicate the attributes like p(arallel) and fuse, and subscripts indicate the branch number.}

\subsection{Feature Extraction} \label{sec:feature-extraction}
As claimed in \cite{kirichenko2020normalizing}, the normalizing flows trained on raw RGB images learn latent representations largely based on local pixel correlations. Such flow models simultaneously increase likelihood for both normal and anomalous images, which defames the distinguishing effectiveness of anomaly detection. 
Moreover, the extracted features imbued with the prior knowledge of the large-scale dataset, such as ImageNet \cite{deng2009imagenet}, are more representative than the simple RGB values in the raw images.
Therefore, we propagate the feature maps extracted by the pre-trained feature extractors \cite{he2016resnet, zagoruyko2016wide} into the flow model.

Since the sizes of anomalies are varied dramatically, we extract multi-scale feature maps as the inputs of the flow model to generalize the anomaly size variation.
Existing detection-only flow-based detection-only methods \cite{rudolph2021differnet,rudolph2022csflow} extract the multi-scale image featurized pyramid upon the image pyramid by an identical extractor. Although the image featurized pyramid involves multi-scale perceptions, all feature maps share the same perceptive field and semantic.
Different from them, we only forward the feature extractor on input images once and sample different stages' last feature maps to construct our multi-scale feature pyramid. The sampled feature maps of these stages are denoted as $\{\boldsymbol{h}_i \in \mathbb{R}^{D_i \times H_i \times W_i}\}_{i=1}^L$, where $L$ is the number of the sample stage, and $L$ is set $3$ like other common methods \cite{gudovskiy2022cflow,rudolph2022csflow}.
\roundb{Consequently, we can formulate our multi-scale feature extraction as follows:}
\begin{equation}
  (\boldsymbol{h}_1, \boldsymbol{h}_2, \boldsymbol{h}_3) = E(\boldsymbol{x}),
\end{equation}
where $E(\cdot)$ denotes the \roundb{pre-trained feature extractor following multi-stage architecture design}.

Notably, we use the feature maps of the first three stages $\{\boldsymbol{h}_1, \boldsymbol{h}_2, \boldsymbol{h}_3\}$ rather than those of the last three \cite{gudovskiy2022cflow} to build our feature pyramid.
Compared with the feature maps of the last $3$ stages with high-level semantic information, the first $3$ stages' feature maps contain more spatial details. 
No matter for the pixel-wise anomaly localization that naturally focuses on spatial structure instead of semantic comprehension or the image-wise anomaly detection that highly depends on the local anomalous regions, the feature maps with abundant details are more applicable.

The low-level feature maps with high resolutions preserve structural details while introducing a high computational cost for the following flow model. To handle the heavy computational burden, we simply downsample them through the average pooling with kernel size 3 and stride 2 to build our feature pyramid $\{\boldsymbol{y}_1,\boldsymbol{y}_2,\boldsymbol{y}_3\}$ as follows:
\begin{equation}
 \boldsymbol{y}_i = \textrm{AvgPooling}(\boldsymbol{h}_i), i = 1,2,3,
\end{equation}
where $\boldsymbol{y}_i \in \mathbb{R}^{D_i \times \frac{H_i}{2} \times \frac{W_i}{2}}$.
Moreover, the weights of the pre-trained feature extractor are frozen and not updated during the training schedule.

\subsection{Multi-scale Normalizing Flows}
The proposed multi-scale flow model in our MSFlow is composed of two parts: 
1) asymmetrical parallel flows for each feature map in the feature pyramid built as Section \ref{sec:feature-extraction} to encode their intrinsic properties;
2) a fusion flow to exchange the global and detailed information of multi-scale feature pyramid.

In the first part, each branch of parallel flows $f_{i}^\textrm{p}$ comprising $k_{i}$ flow blocks respectively encode the extracted feature map $\boldsymbol{y}_i$ and transform it to the latent feature map $\boldsymbol{z}_i^\textrm{p}=f_{i}^\textrm{p}(\boldsymbol{y}_i)$. 
In particular, our parallel flows adopt an asymmetrical architecture as shown in Fig. \ref{fig:framework}, where $k_{i+1} > k_{i}$.
Our parallel flows build the st-network $st_\textrm{p}(\cdot )$ with the $3\times3$ convolutions, which automatically capture the spatial contextual information neglected in \cite{gudovskiy2022cflow}.
For details, the $st_\textrm{p}(\cdot )$ networks in all parallel flows share the same architecture:
\rounda{\begin{equation} \label{equation:st-network}
  st_\textrm{p}(\cdot ) = 
  \textrm{Conv3} \circ \textrm{ReLU} \circ \textrm{LN} \circ \textrm{Conv3} (\cdot ),
\end{equation}}
where Conv3 refers to the $3 \times 3$ convolution, ReLU is the ReLU non-linear activation, and LN denotes the layer normalization \cite{ba2016layer}.
\rounda{
Particularly, the hidden dimension between two $3 \times 3$ convolutions is set identical to the input dimension of the first one $3 \times 3$ convolution for simplicity.}
The additional layer normalization not only stabilizes the optimization but also improves the performance of anomaly detection with the statistic on anomaly-free training data \cite{song2019unsupervised}.
Besides, positional encodings \cite{vaswani2017attention} are inserted into parallel flow blocks as the condition. Although the CNN layers capture the global perception, they fail to perceive the absolute position of each feature vector. For anomaly detection, especially industrial defect detection, where all industrial parts share the same photography pose, the perception of the absolute position is conducive to this specific task.

\begin{figure}
  \centering
  \includegraphics[width=0.85\linewidth]{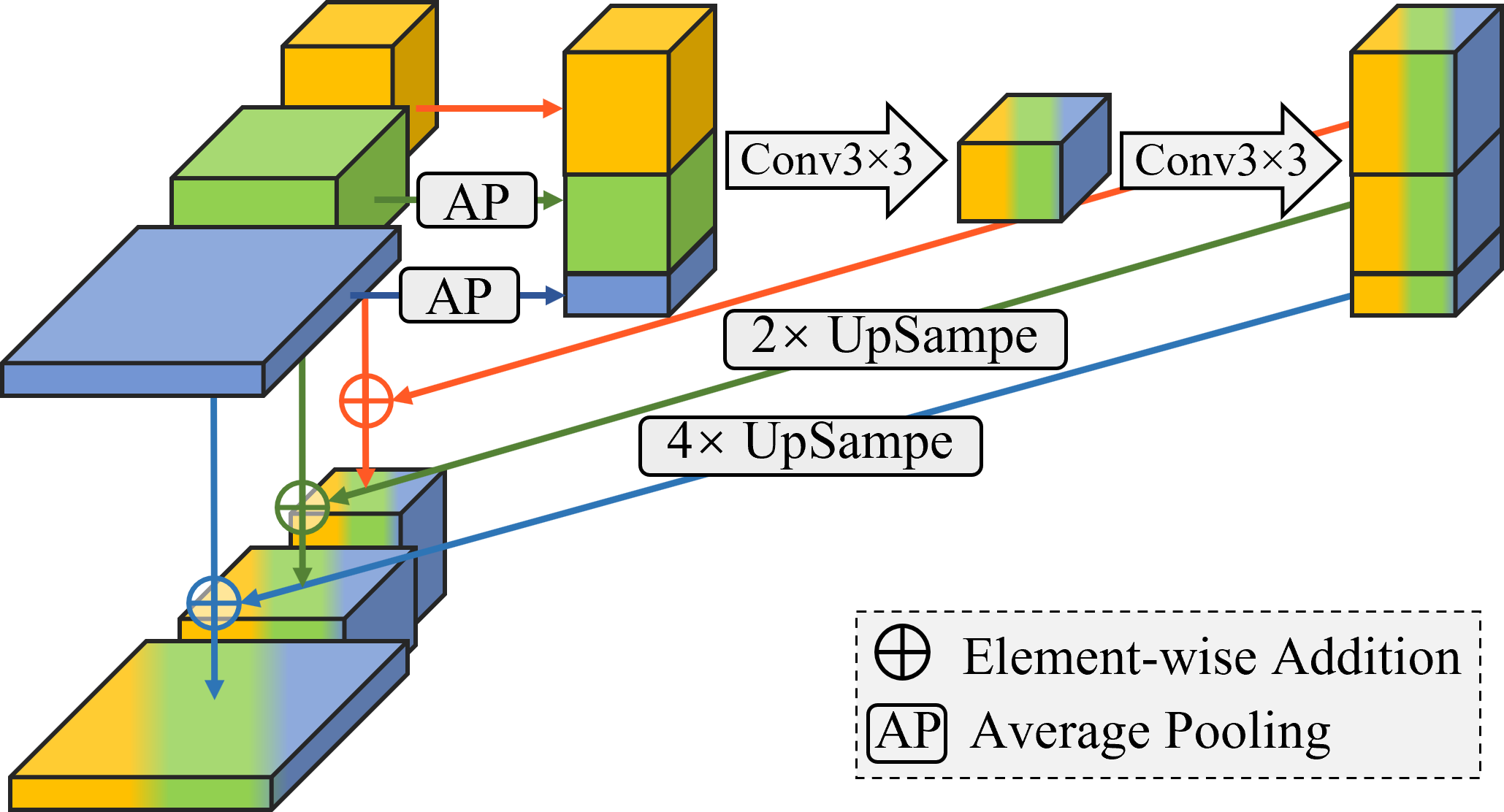}
  \caption{The detailed architecture of the fusion network $g^\textrm{fuse}$ in st-network of the fusion flow. $AP$ denotes the average pooling.} \label{fig:fuse-network}
  \label{fig:fusion-network}
\end{figure}

\rounda{
A fusion flow $f_\textrm{fuse}$ is further attached to the encoded feature pyramid
$\{\boldsymbol{z}_i^\textrm{p}\}_{i=1}^3$
and estimates the final multi-scale outputs
$\{\boldsymbol{z}_i^\textrm{fuse}\}_{i=1}^3$.
The proposed fusion flow $f_\textrm{fuse}$ efficiently exploits multi-scale perception by exchanging information captured on different scales.
Unlike the common flow blocks \cite{dinh2016realnvp,kingma2018glow} that only process on a single feature map, our fusion flow estimates the multi-scale feature maps simultaneously and fuses their information. 
Particularly, a fusion network $g^\textrm{fuse}$ first fuses the multi-scale inputs $\{\boldsymbol{a}_i\}_{i=1}^3$ and outputs fused feature maps $\{\boldsymbol{a}_i^\textrm{fuse}\}_{i=1}^3$ before learning the $scale$ weights $\boldsymbol{w}^\textrm{s}_i$ and $shift$ weights $\boldsymbol{w}^\textrm{t}_i$ in the st-network of our fusion flow. Such an st-network is formulated as follows:
\begin{equation}
  \begin{aligned}
    (\boldsymbol{a}_1^\textrm{fuse}&, \boldsymbol{a}_2^\textrm{fuse}, \boldsymbol{a}_3^\textrm{fuse}) = g^\textrm{fuse}(\boldsymbol{a}_1, \boldsymbol{a}_2, \boldsymbol{a}_3); \\
    \boldsymbol{w}^\textrm{s}_i &= g_i^\textrm{s}(\boldsymbol{a}_i^\textrm{fuse}), \textrm{where}\, i =1, 2, 3; \\
    \boldsymbol{w}^\textrm{t}_i &= g_i^\textrm{t}(\boldsymbol{a}_i^\textrm{fuse}), \textrm{where}\, i =1, 2, 3.
  \end{aligned}
\end{equation}
Inspired by the lightweight block \cite{yu2021lite} for high-resolution pose estimation, the fusion network $g^\textrm{fuse}$ in our fusion flow $f_\textrm{fuse}$ is designed as illustrated in Fig. \ref{fig:fuse-network}.
The fully-convolutional $g^\textrm{fuse}$ provides a bridge for multi-scale feature maps to exchange information across all scales. 
The $g^\textrm{fuse}$ first shrinks all multi-scale inputs $\{\boldsymbol{a}_1,\boldsymbol{a}_2,\boldsymbol{a}_3\}$ to the minimum size of them (size of $\boldsymbol{a}_3$) by the average pooling and concatenate them together. The concatenated features are fused by two convolutional layers with a similar structure as Eq. \ref{equation:st-network}, where the middle channels are narrowed by dividing 4. The fused single-scale feature map is split along the channel dimension and rescaled to multi-scale again. Finally, the fused multi-scale feature maps are added with their respective inputs to get the fused outputs $\{\boldsymbol{x}_i^\textrm{fuse}\}_{i=1}^3$.
Based on the fused feature pyramid that receives the multi-scale perception, $3$ pairs of $3\times3$ convolutions $g_i^\textrm{s}$ and $g_i^\textrm{t}$ independently learn the \textit{scale} and \textit{shift} weights $\boldsymbol{w}^\textrm{s}_i$ and $\boldsymbol{w}^\textrm{t}_i$ for each scale.
}

\subsection{Learning Objective}
As introduced in Section \ref{sec:theory-background}, the optimization target of the normalizing flows is uniform and simple. Because we employ the feature extracted maps $\{\boldsymbol{y}_i\}_{i=1}^{3}$ as the input instead of the raw RGB image $\boldsymbol{x}$, the optimization target shifts to the maximization of the likelihood $p_Y(\boldsymbol{y})$ in the feature space $Y$.
The proposed MSFlow is composed of several flow submodels: 3 parallel flows $\{f_{i}^\textrm{p}\}_{i=1}^3$ and one fusion flow $f_\textrm{fuse}$, which are summarized as follows:
\begin{equation}
  \begin{aligned}
    \boldsymbol{z}_i^\textrm{p}&=f_{i}^\textrm{p}(\boldsymbol{y}_i), \, \textrm{where} \,\, i=1,2,3;\\ 
    (\boldsymbol{z}_1^\textrm{fuse}&,\boldsymbol{z}_2^\textrm{fuse}, \boldsymbol{z}_3^\textrm{fuse}) = f_\textrm{fuse}(\boldsymbol{z}_1, \boldsymbol{z}_2, \boldsymbol{z}_3).
  \end{aligned}
\end{equation}
Accordingly, there are 4 Jacobian determinants in our model: $\{ \left\lvert \textrm{det}J_{{f_{i}^\textrm{p}}^{-1}}\right\rvert \}_{i=1}^3$ and $\left\lvert \textrm{det}J_{{f_\textrm{fuse}}^{-1}} \right\rvert$. 
\rounda{Correspondingly, the global learning objective $\mathcal{L}$ can be formulated as the minimization of the negative log-likelihood $-\textrm{log}p_Y(\boldsymbol{y})$:
\begin{align} 
    \mathcal{L} = & -\textrm{log}\,p_Y(\boldsymbol{y})  \\
    & \quad \,\,\,\, \frac{\left\lVert \boldsymbol{z} \right\rVert_2^2}{2}  \quad \,\,\,- \, \quad  \qquad  \quad   \textrm{log}\left\lvert \textrm{det}J_{f^{-1}} \right\rvert  \notag \\
    = &\overbrace{\sum_{i = 1}^{3} \frac{\left\lVert \boldsymbol{z}_i^\textrm{fuse} \right\rVert_2^2}{2}}
    - \overbrace{\left[  \sum_{i = 1}^{3} \textrm{log}\left\lvert \textrm{det}J_{{f_{i}^\textrm{p}}^{-1}} \right\rvert 
    + \textrm{log}\left\lvert \textrm{det}J_{f_\textrm{fuse}^{-1}}\right\rvert  \right]}, \notag \label{equation:global-loss}
\end{align}
where $f$ refers to the whole flow model containing all parallel flows $\{f_{i}^\textrm{p}\}_{i=1}^3$ and the fusion flow $f_\textrm{fuse}$. 
From the global perspective, our learning objective $\mathcal{L}$ is consistent with the Eq. \ref{equation:loss-function}. }

\subsection{Anomaly Score Calculations} \label{sec:aggregation}
After training on the anomaly-free images, our MSFlow can transform the feature distribution of normal data to the base distribution. According to the Eq. \ref{equation:loss-function} the pixel-wise log-likelihood $\textrm{log} \,\hat{p}_Y (\boldsymbol{y}_{ijk})$ located at $(j, k)$ on the $i$-th feature map $\boldsymbol{y}_i$ can be estimated as:
\begin{equation}
  \textrm{log}\,\hat{p}_Y (\boldsymbol{y}_{ijk}) = -\frac{\left\lVert \boldsymbol{z}_{ijk} \right\rVert_2^2}{2} + \textrm{log}\left\lvert \textrm{det}J_{f^{-1}}(\boldsymbol{y}_{ijk})\right\rvert.
\end{equation}
Unlike \cite{gudovskiy2022cflow} which independently estimates the likelihood of the features $\boldsymbol{y}_{ijk}$ at each position, we globally estimate the entire feature map's likelihood and fail to calculate the pixel-wise Jacobian determinant $\left\lvert \textrm{det}J_{f^{-1}}(\boldsymbol{y}_{ijk}) \right\rvert$. Instead, we directly compute the anomaly score based on the log-likelihood $\textrm{log}\,\hat{p}_Z (\boldsymbol{z}_{ijk})= -\frac{\left\lVert \boldsymbol{z}_{ijk} \right\rVert_2^2}{2}$, which is the one-to-one mapping of $\textrm{log}\,\hat{p}_Y (\boldsymbol{y}_{ijk})$.

Before aggregation, the multi-scale log-likelihood maps are first upsampled to the input image size $(H_{img}, W_{img})$ by the bilinear interpolating. Different aggregation strategies are employed for pixel-wise anomaly localization and image-wise anomaly detection. For the pixel-wise anomaly localization, the rescaled log-likelihood maps $\{\textrm{log}\,\hat{P}_i\}_{i=1}^3$ are converted to multi-level probability maps $\{\hat{P}_i = e^{\textrm{log}\,\hat{P}_i} \}_{i=1}^3$, which are further aggregated through the element-wise addition as follows:
\rounda{
\begin{equation}
  P_{add} = \sum_{i = 1}^{3} e^{\textrm{log}\,\hat{P}_i} = \sum_{i = 1}^{3} \hat{P}_i.
\end{equation}}
Differently, the rescaled log-likelihood maps $\{\textrm{log}\,\hat{P}_i\}_{i=1}^3$ are summed first and then converted to the probability map $P_{mul}$ for the image-wise anomaly detection. It can also be viewed as the multiplication of multi-level probability maps $\{\hat{P}_i \}_{i=1}^3$ as follows:
\rounda{
\begin{equation}
  P_{mul} = e^{\sum_{i = 1}^{3} \textrm{log}\,\hat{P}_i} = \prod_{i = 1}^{3} e^{\textrm{log}\,\hat{P}_i} = \prod_{i = 1}^{3} \hat{P}_i.
\end{equation}}
The multiplication that suppresses the noise of a certain layer is more suitable for the global anomaly score calculation. However, pixel-wise anomaly localization benefits from the addition aggregation that preserves multi-scale information for various scale anomalies.
The final pixel-wise anomaly score map $S_{loc}$ and the image-wise anomaly score $s_{det}$ are respectively calculated as follows:
\begin{align}
  S_{loc} &= \textrm{max}(P_{add}) - P_{add}; \\
  S_{mul} &= \textrm{max}(P_{mul})-P_{mul}, \notag \\
  s_{det} &= \frac{1}{K} \sum_{1}^K \textrm{top}K(S_{mul}),
\end{align}
where the max$(\cdot)$ and top$K(\cdot)$ respectively sample the maximum and the largest $K$ values.
The previous methods either take the maximum \cite{gudovskiy2022cflow, zavrtanik2021riad} or the mean \cite{rudolph2022csflow,rudolph2021differnet} in the pixel-wise anomaly score map as the global anomaly score. The former as an image-wise anomaly score is too sensitive to the noise, the latter is robust while imperceptible to small anomalous regions. 
To compensate for their individual drawbacks, we combine them together and take the mean of the top$K$ values as the global image-wise anomaly score.
\rounda{
Different to DevNet \cite{pang2019devnet} sampling top$K$ feature values along channel dimension, our MSFlow takes average on the top$K$ pixel-wise anomaly in the spatial dimensions.}
The maximum and mean are the special cases of our solution when $K$ is set to $1$ or $H_{img} \times W_{img}$.

\begin{table*}[t] 
  \centering
  \caption{The comparisons of image-wise anomaly detection performance (AUCROC in \%) on the MVTec AD benchmark. The best results are highlighted in bold, and the second-best results are underlined.}\label{tab:det-comparison-mvtec}
  \resizebox{\textwidth}{!}{
  \begin{tabular}{cl|cccc|c|cccc}
  \myrule
  \multicolumn{2}{c|}{Taxonomy}                                 & \multicolumn{4}{c|}{Reconstruction-based}                  & Clustering-based & \multicolumn{4}{c}{Flow-based}                            \\ \hline
  \multicolumn{2}{c|}{Method}                                   & CutPaste \cite{li2021cutpaste}     & DRAEM \cite{zavrtanik2021draem}       & SSPCAB \cite{ristea2021self}       & RD4AD \cite{deng2022distillation}    & PatchCore \cite{roth2021patchcore}      & DifferNet \cite{rudolph2021differnet}  & CFlow-AD \cite{gudovskiy2022cflow}     & CSFlow \cite{rudolph2022csflow}      & \textbf{MSFlow} \\ \hline
  \multicolumn{2}{c|}{Venue}                                    & CVPR'21      & CVPR'21      & CVPR'22       & CVPR'22       & CVPR'22         & WACV'21    & WACV'22       & WACV'22      & \textbf{Ours} \\ \hline
  \multicolumn{1}{c|}{\multirow{6}{*}{\rotatebox{90}{Texture}}} & Carpet       & 93.9         & 97.0         & 98.2          & 98.9          & 98.7            & 92.9       & {\ul 99.3}    & \textbf{100} & \textbf{100}  \\
  \multicolumn{1}{c|}{}                          & Grid         & \textbf{100} & {\ul 99.9}   & \textbf{100}  & \textbf{100}  & 98.2            & 84.0       & 99.0          & 99.0         & 99.8          \\
  \multicolumn{1}{c|}{}                          & Leather      & \textbf{100} & \textbf{100} & \textbf{100}  & \textbf{100}  & \textbf{100}    & 97.1       & {\ul 99.7}    & \textbf{100} & \textbf{100}  \\
  \multicolumn{1}{c|}{}                          & Tile         & 94.6         & {\ul 99.6}   & \textbf{100}  & 99.3          & 98.7            & 99.4       & 98.0          & \textbf{100} & \textbf{100}  \\
  \multicolumn{1}{c|}{}                          & Wood         & 99.1         & 99.1         & 99.5          & 99.2          & 99.2            & {\ul 99.8} & 96.7          & \textbf{100} & \textbf{100}  \\ \cline{2-11} 
  \multicolumn{1}{c|}{}                          & \tb{Average} & 97.5         & 99.1         & 99.5          & 99.5          & 99.0            & 94.6       & 98.5          & {\ul 99.8}   & \textbf{100}  \\ \hline
  \multicolumn{1}{c|}{\multirow{11}{*}{\rotatebox{90}{Object}}} & Bottle       & 98.2         & 99.2         & 98.4          & \textbf{100}  & \textbf{100}    & 99.0       & 99.0          & {\ul 99.8}   & \textbf{100}  \\
  \multicolumn{1}{c|}{}                          & Cable        & 81.2         & 91.8         & 96.9          & 95.0          & \textbf{99.5}   & 95.9       & 97.6          & {\ul 99.1}   & \textbf{99.5} \\
  \multicolumn{1}{c|}{}                          & Capsule      & 98.2         & 98.5         & \textbf{99.3} & 96.3          & 98.1            & 86.9       & 99.0          & 97.1         & {\ul 99.2}    \\
  \multicolumn{1}{c|}{}                          & Hazelnut     & 98.3         & \textbf{100} & \textbf{100}  & {\ul 99.9}    & \textbf{100}    & 99.3       & 98.9          & 99.6         & \textbf{100}  \\
  \multicolumn{1}{c|}{}                          & Metal Nut    & {\ul 99.9}   & 98.7         & \textbf{100}  & \textbf{100}  & \textbf{100}    & 96.1       & 98.6          & 99.1         & \textbf{100}  \\
  \multicolumn{1}{c|}{}                          & Pill         & 94.9         & 98.9         & \textbf{99.8} & 96.6          & 96.6            & 88.8       & 99.0          & 98.6         & {\ul 99.6}    \\
  \multicolumn{1}{c|}{}                          & Screw        & 88.7         & 93.9         & {\ul 97.9}    & 97.0          & 98.1            & 96.3       & \textbf{98.9} & 97.6         & 97.8          \\
  \multicolumn{1}{c|}{}                          & Toothbrush   & 99.4         & \textbf{100} & \textbf{100}  & {\ul 99.5}    & \textbf{100}    & 98.6       & 98.9          & 91.9         & \textbf{100}  \\
  \multicolumn{1}{c|}{}                          & Transistor   & 96.1         & 93.1         & 92.9          & 96.7          & \textbf{100}    & 91.1       & 93.3          & {\ul 99.3}   & \textbf{100}  \\
  \multicolumn{1}{c|}{}                          & Zipper       & 99.9         & \textbf{100} & \textbf{100}  & 98.5          & 98.8            & 95.1       & 99.1          & {\ul 99.7}   & \textbf{100}  \\ \cline{2-11} 
  \multicolumn{1}{c|}{}                          & \tb{Average} & 95.5         & 97.4         & 98.5          & 98.0          & {\ul 99.1}      & 94.7       & 98.2          & 98.2         & \textbf{99.6} \\ \hline
  \multicolumn{2}{c|}{\textbf{Total Average}}                   & 96.1         & 98.0         & 98.9          & 98.5          & {\ul 99.1}      & 94.7       & 98.3          & 98.7         & \textbf{99.7} \\ \myrule
  \end{tabular}}
  \end{table*}

\section{Experiment} \label{sec:experiment}
We evaluate the MSFlow on the two image anomaly detection datasets as well as one video violation detection dataset and present the comparisons with existing methods. We also perform comprehensive ablation studies to investigate the effectiveness of each proposed component in our MSFlow.

\subsection{Experimental Settings}
\subsubsection{Datasets}
To highlight the superiority and generality of our method, we mainly conduct performance comparisons with other methods on two image anomaly detection datasets and a video violation detection dataset. For all these datasets, no data augmentation except resizing is applied.

\textit{MVTec Anomaly Detection (MVTec)}\cite{bergmann2019mvtec}: The widely-used MVTec AD benchmark contains 15 industrial product classes, where 10 objects and 5 textures. 
There are total 3,629 images for training and 1,725 images for testing, and the train set size varies and ranges from 60 for \textit{toothbrush} to 391 for \textit{screw} for each class. 
Only defect-free samples are included in the train set, the test set includes both defect-free samples and anomalous samples of various defective types (up to 8 for \textit{cable}). 
The images in the MVTec AD benchmark are high-quality with resolutions from $700\times700$ to $1024 \time 1024$. We resize images of all classes except the \textit{transistor} to $512\times512$, and $256 \times 256$ for \textit{transistor} whose defective size of misplaced samples is large up to 42\% average.

\textit{Magnetic Tile Defects (MTD)} \cite{huang2020mtd}:
The MTD dataset is specialized for \textit{tile} with 952 defect-free images and 382 defective images. For a fair comparison, we use 80\% defect-free images for training and the rest are used for testing along with all defective images. In particular, the images in the MTD dataset are resized to $192\times192$ since the images are low-resolution.

\textit{mini Shanghai Tech Campus (mSTC)}:
Besides image anomaly detection datasets, we also extend our method to the video violation detection dataset mSTC \cite{defard2021padim, roth2021patchcore}. The mSTC dataset comprises every fifth frame sampled from the training and test videos in the standard STC dataset \cite{luo2017stc}. The videos for training only contain normal behaviors while the test set includes diverse violations such as chasing and brawling. All sampled frames are resized to $256\times384$ during both training and testing.

\subsubsection{Evaluation Metrics}
Following \cite{bergmann2019mvtec}, the performance of both image-wise anomaly detection and pixel-wise anomaly localization are evaluated through the area under the receiver-operator curve (AUCROC) for each class. Furthermore, we compute the class-average AUCROC score as the result on the whole dataset for the multi-class datasets: the MVTec AD and mSTC datasets. Particularly on the challenging MVTec AD benchmark, we also measure the per-region-overlap (PRO) score \cite{bergmann2020uninformed} for pixel-wise anomaly localization. 
\rounda{The PRO curve is calculated as the overlap between each connected region within the ground truth mask. The PRO score is the area under the PRO curve, when an average per-pixel false-positive rate of $30\%$ for the entire dataset is reached.}
The PRO score treats the defects with different sizes equally and attaches importance to the connectivity of localization results, which can be viewed as the region-wise AUCROC score.

\subsubsection{Implementation Details}
For a fair comparison with existing methods, we utilize the most widely-used WideResNet-50 (WRN-50) \cite{zagoruyko2016wide} and ResNet-18 \cite{he2016resnet} as the feature extractors.
Before feeding into our parallel flows, the activation outputs of the first 3 stages in these two feature extractors are downsampled by the average pooling for efficiency.
Our parallel flows cascade more flow blocks for the higher-level feature map with more channels. Specifically, 2, 5, and 8 fully-convolutional flow blocks are respectively stacked for the feature maps of stage1, stage2, and stage3. In all flow blocks of these three parallel flows, the 2D positional encodings with channels=64 are inserted to capture the absolute position. For information exchange, a fusion flow is attached to fuse the outputs of our three parallel flows. 
Our flow model is trained from scratch on one 2080Ti GPU card with a batch size of 16. For optimization, we adopt the Adam optimizer with an initial learning rate of $1e^{-4}$, and the learning rate is dropped by $3$ at 70\% and 90\% of the whole training schedule. We set different training epochs for these 3 industrial anomaly detection datasets according to the amount of training data: 100 epochs for MVTec AD and 30 epochs for MTD. As for the large-scale mSTC, the MSFlow is only trained for 10 epochs. 
Our MSFlow is training-time-friendly, costing less than half the training schedule of the previous methods \cite{gudovskiy2022cflow,rudolph2022csflow,rudolph2021differnet} yet achieving superior performance.

\begin{table*}[t] 
  \centering
  \caption{The comparisons of pixel-wise anomaly localization performance (AUCROC and RPO in \%) on the MVTec AD benchmark. The best results are highlighted in bold, and the second-best results are underlined. The PRO scores are not evaluated in the DRAEM\cite{zavrtanik2021draem} and SSPCAB\cite{ristea2021self}, replaced with `-'.}\label{tab:loc-comparison-mvtec}
  \begin{tabular}{cl|ccc|ccc|cc}
  \myrule
  \multicolumn{2}{c|}{Taxonomy}                               & \multicolumn{3}{c|}{Reconstruction-based} & \multicolumn{3}{c|}{Clustering-based}   & \multicolumn{2}{c}{Flow-based} \\ \hline
  \multicolumn{2}{c|}{Method}                                 & DRAEM \cite{zavrtanik2021draem}         & SSPCAB \cite{ristea2021self}        & RD4AD   \cite{deng2022distillation}               & SPADE \cite{cohen2020spade}            & PaDiM \cite{defard2021padim}      & PatchCore \cite{roth2021patchcore}        & CFlow-AD \cite{gudovskiy2022cflow}               & \tb{MSFlow}  \\ \hline
  \multicolumn{2}{c|}{Venue}                                  & CVPR'21        & CVPR'22        & CVPR'22                & ArXiv'21          & ICPR'21     & CVPR'22           & WACV'22                 & \tb{Ours}  \\ \hline
  \multicolumn{1}{c|}{\multirow{6}{*}{\rotatebox{90}{Texture}}} & Carpet     & 95.5 / -       & 95.0 / -       & 98.9 / 97.0            & 97.5 / 94.7       & 99.1 / 96.2 & 99.1 / 96.6       & {\ul 99.3} / {\ul 97.7} & \tb{99.4} / \tb{99.6}   \\
  \multicolumn{1}{c|}{}                          & Grid       & \tb{99.7} / -  & 99.5 / -       & 99.3 / {\ul 97.6}      & 93.7 / 86.7       & 97.3 / 94.6 & 98.7 / 95.9       & 99.0 / 96.1             & {\ul 99.4} / \tb{99.1}  \\
  \multicolumn{1}{c|}{}                          & Leather    & 98.6 / -       & {\ul 99.5} / - & 99.4 / 99.1            & 97.6 / 97.2       & 99.2 / 97.8 & 99.3 / 98.9       & \tb{99.7} / {\ul 99.4}  & \tb{99.7} / \tb{99.9}   \\
  \multicolumn{1}{c|}{}                          & Tile       & {\ul 99.2} / - & \tb{99.3} / -  & 95.6 / 90.6            & 87.4 / 75.9       & 94.1 / 86.0 & 95.9 / 87.4       & 98.0 / {\ul 94.3}       & 98.2 / \tb{95.3}        \\
  \multicolumn{1}{c|}{}                          & Wood       & 96.4 / -       & {\ul 96.8} / - & 95.3 / 90.9            & 88.5 / 87.4       & 94.9 / 91.1 & 95.1 / 89.6       & 96.7 / {\ul 95.8}       & \tb{97.1} / \tb{96.6}   \\ \cline{2-10}      
  \multicolumn{1}{c|}{}                          & Average    & 97.9 / -       & 98.0 / -       & 97.7 / 95.0            & 92.9 / 88.4       & 96.9 / 93.1 & 97.6 / 93.7       & {\ul 98.5} / {\ul 96.7} & \tb{98.8} / \tb{98.1}   \\ \hline
  \multicolumn{1}{c|}{\multirow{11}{*}{\rotatebox{90}{Texture}}} & Bottle     & \tb{99.1} / -  & 98.8 / -       & 98.7 / 96.6            & 98.4 / 95.5       & 98.3 / 94.8 & 98.6 / 96.1       & {\ul 99.0} / {\ul 96.8} & {\ul 99.0} / \tb{98.5}  \\
  \multicolumn{1}{c|}{}                          & Cable      & 94.7 / -       & 96.0 / -       & 97.4 / 91.0            & 97.2 / 90.9       & 96.7 / 88.8 & \tb{98.5} / 92.6  & {\ul 97.6} / {\ul 93.5} & \tb{98.5} / \tb{93.7}   \\
  \multicolumn{1}{c|}{}                          & Capsule    & 94.3 / -       & 93.1 / -       & 98.7 / {\ul 95.8}      & {\ul 99.0} / 93.7 & 98.5 / 93.5 & 98.9 / 95.5       & 99.0 / 93.4             & \tb{99.1} / \tb{98.4}   \\
  \multicolumn{1}{c|}{}                          & Hazelnut   & {\ul 99.7} / - & \tb{99.8} / -  & 98.9 / 95.5            & 99.1 / 95.4       & 98.2 / 92.6 & 98.7 / 93.9       & 98.9 / \tb{96.7}        & 98.7 / {\ul 96.6}       \\
  \multicolumn{1}{c|}{}                          & Metal Nut  & \tb{99.5} / -  & 98.9 / -       & 97.3 / 92.3            & 98.1 / {\ul 94.4} & 97.2 / 85.6 & 98.4 / 91.3       & 98.6 / 91.7             & {\ul 99.3} / \tb{97.6}  \\
  \multicolumn{1}{c|}{}                          & Pill       & 97.6 / -       & 97.5 / -       & 98.2 / \tb{96.4}       & 96.5 / 94.6       & 95.7 / 92.7 & 97.6 / 94.1       & \tb{99.0} / 95.4        & {\ul 98.8} / {\ul 96.0} \\
  \multicolumn{1}{c|}{}                          & Screw      & 97.6 / -       & \tb{99.8} / -  & {\ul 99.6} / \tb{98.2} & 98.9 / 96.0       & 98.5 / 94.4 & 99.4 / {\ul 97.9} & 98.9 / 95.3             & 99.1 / 94.2             \\
  \multicolumn{1}{c|}{}                          & Toothbrush & 98.1 / -       & 98.1 / -       & \tb{99.1} / {\ul 94.5} & 97.9 / 93.5       & 98.8 / 93.1 & 98.7 / 91.4       & {\ul 98.9} / \tb{95.1}  & 98.5 / 91.6             \\
  \multicolumn{1}{c|}{}                          & Transistor & 90.9 / -       & 87.0 / -       & 92.5 / 78.0            & 94.1 / {\ul 87.4} & 97.5 / 84.5 & 96.4 / 83.5       & {\ul 98.0} / 81.4       & \tb{98.3} / \tb{99.8}   \\
  \multicolumn{1}{c|}{}                          & Zipper     & 98.8 / -       & 99.0 / -       & 98.2 / 95.4            & 96.5 / 92.6       & 98.5 / 95.9 & 98.9 / {\ul97.1}  & {\ul 99.1} / 96.6       & \tb{99.2} / \tb{99.4}   \\ \cline{2-10}      
  \multicolumn{1}{c|}{}                          & Average    & 97.0 / -       & 96.8 / -       & 97.9 / 93.4            & 97.6 / 93.4       & 97.8 / 91.6 & 98.4 / 93.3       & {\ul 98.7} / {\ul 93.6} & \tb{98.8} / \tb{96.6}   \\ \hline
  \multicolumn{2}{c|}{Total Average}                          & 97.3 / -       & 97.2 / -       & 97.8 / 93.9            & 96.0 / 91.7       & 97.5 / 92.1 & 98.1 / 93.5       & {\ul 98.6} / {\ul 94.6} & \tb{98.8} / \tb{97.1}   \\ \myrule
  \end{tabular}
  \end{table*}

\subsection{Comparisons with Other Methods}
\subsubsection{MVTec AD}
We compare our MSFlow with the previous reconstruction-based methods \cite{zavrtanik2021draem, cohen2020spade,li2021cutpaste,deng2022distillation}, clustering-based methods \cite{cohen2020spade,defard2021padim,roth2021patchcore} and flow-based methods \cite{rudolph2021differnet, gudovskiy2022cflow,rudolph2022csflow} on the MVTec AD benchmark. Our method exhibits consistently superior performance regardless of the image-wise anomaly detection or the pixel-wise anomaly localization.

The image-wise anomaly detection comparisons are presented in Table \ref{tab:det-comparison-mvtec}. Since only the class average AUCROC scores are provided in  
Compared with prior methods of all three categories, our MSFlow achieves the best or the second-best detection performance in all classes except the \textit{grid} and the \textit{screw}, whose accuracies are only 0.1\% lower than the second-best performance. The reconstruction-based methods achieve the best detection accuracy on some classes such as \textit{grid} and \textit{pill}. However, their detection AUCROC scores are inferior on other classes like \textit{transistor} including misplaced defects. The near-perfect detection AUCROC scores on all classes highlight the effectiveness and generality of anomaly size variation of the proposed MSFlow.
Naturally, our MSFlow also achieves the best class-average detection performance on either \textit{texture} classes or \textit{object} classes with 100\% and 99.6\% AUCROC scores. The overall class-average detection accuracy of the MSFlow is up to 99.7\%, which is 0.6\% higher than that of the previous SOTA \textit{PatchCore} with the same feature extractor (WRN-50). The proposed MSFlow even outperforms the ensemble version of PatchCore (99.6\%) with large feature extractors including WRN-101 \cite{he2016resnet}, DenseNet-201\cite{huang2017densenet} and ResNext-101 \cite{xie2017resnext}. 

The performance comparisons of pixel-wise anomaly localization with prior methods are shown in Table \ref{tab:loc-comparison-mvtec}. Among flow-based methods, DifferNet\cite{rudolph2021differnet} and CSFlow\cite{rudolph2022csflow} are designed for image-wise anomaly detection without localization, hence only CFlow-AD\cite{gudovskiy2022cflow} is compared with our MSFlow in anomaly localization.
The proposed MSFlow achieves high localization accuracies in all industrial classes, summarized in the class-average AUCROC score of 98.8\% and PRO score of 97.1\%. 
The PRO scores of our MSFlow are overwhelmingly superior to other methods, especially on the \textit{carpet, grid, leather, transistor, and zipper} with higher than 99\% PRO \rounda{score}. The overall RPO score is dramatically improved by 2.5\%, which demonstrates the high-precision localization and noise suppression capability of the proposed MSFlow.

\begin{table}[t]
  \caption{The comparisons of image-wise anomaly detection performance (AUCROC in \%) on the MTD dataset. The best results are highlighted in bold, and the second-best results are underlined.} \label{tab:det-comparison-mtd}
  \centering
  \resizebox{\linewidth}{!}{
  \begin{tabular}{c|c|c|c}
  \myrule
  Method             & GANomaly \cite{akcay2018ganomaly}  & DifferNet \cite{rudolph2021differnet}    & PaDiM  \cite{defard2021padim}       \\ \hline
  Det. AUCROC  & 76.6                               & 97.7                                     & 98.7          \\ \hline \hline
  Method             & PatchCore \cite{roth2021patchcore} & CSFlow  \cite{rudolph2022csflow}         & \textbf{MSFlow(Ours)} \\ \hline
  Det. AUCROC  & 97.9                               & \textbf{99.3}                            & {\ul 99.2}    \\ \myrule
  \end{tabular}}
  \end{table}

\begin{table}[t] 
  \caption{The comparisons of pixel-wise anomaly localization performance (AUCROC in \%) on the mSTC dataset. The best results are highlighted in bold, and the second-best results are underlined. } \label{tab:loc-comparison-mstc}
  \centering
  \begin{tabular}{c|c|c|c|c}
  \myrule                
  Method                     & \makecell{SPADE \\ \cite{cohen2020spade}}  & \makecell{PaDiM \\ \cite{defard2021padim}} & \makecell{PatchCore \\ \cite{roth2021patchcore}}    & \makecell{\textbf{MSFlow} \\ \textbf{(Ours)}} \\ \hline
  Loc. AUCROC         & 89.9                         &  91.2                        & {\ul 91.8}                          & \textbf{93.0}    \\ \myrule
  \end{tabular}
  \end{table}

\subsubsection{Other Datasets}
The anomaly detection performance on MTD and the violation localization performance on mSTC are presented in Table \ref{tab:det-comparison-mtd} and Table \ref{tab:loc-comparison-mstc}, respectively. On the MTD dataset specialized for \textit{Tile}, our MSFlow achieves the second-best detection AUCROC score of 99.2\%, negligibly inferior to the detection-only CSFlow by 0.1\%. On the non-image dataset mSTC where pedestrian violations are treated as anomalies.
Following \cite{roth2021patchcore}, the proposed MSFlow is compared with clustering-based SAPDE\cite{cohen2020spade}, PaDiM\cite{defard2021padim} and PatchCore\cite{roth2021patchcore} on the mSTC dataset. Although the violation detection requires semantic comprehension, our MSFlow achieves the best violation localization AUCROC score of 93.0\%, outperforming the clustering-based PatchCore \cite{roth2021patchcore} by 1.2\%.


\subsection{Ablation Study}
The comprehensive ablation studies are conducted on the MVTec AD benchmark. If not specifically stated, all results are obtained on the MSFlow with WRN-50 as the feature extractor.

\subsubsection{Multi-scale Feature Pyramid Extraction}
The feature extraction is simple but essential for flow-based methods to achieve high anomaly detection performance. We not only explore the effect of feature maps of different levels but also reveal the influence of different feature extractors.

\vspace{3pt}\textbf{The Effect of Feature Maps of Different Levels}. We only change the feature maps fed into the following flow model but keep other implementations identical. The overall average detection AUCROC scores, localization AUCROC scores, and localization PRO scores are presented in Table \ref{as-feature}. To fairly compare the effect of feature maps of stage1 and stage4, the feature pyramid comprising feature maps of stage2 and stage3 is treated as the baseline. With these two feature maps, our MSFlow also achieves a high detection AUCROC score of 99.4\% while performing inferiorly on anomaly localization. When added stage4's feature map, performance even becomes worse in both anomaly detection and localization. The feature map of stage4 contains high-level semantic information but lacks details. With such a feature map as input, the flow model is induced to focus on semantic understanding rather than spatial structure perception. 
Both anomaly detection and localization performance are promoted with the help of the feature map of stage1 as the third row in Table \ref{as-feature}, which verifies the significance of spatial details for anomaly detection and localization. 

\begin{table}[!htb]
  \centering
  \rounda{
  \caption{The ablation studies of different feature maps. S$ijk$ refers to feature maps of stage$i$, stage$j$, and stage$k$. MP and AP denote the max pooling and average pooling, k denotes the kernel size. The pair of GFLOPs respectively denotes the GFLOPs of the feature extractor and the flow model. The best results are highlighted in bold.} \label{as-feature}
  \resizebox{\linewidth}{!}{
  \begin{tabular}{@{}l|c|c|c|c@{}}
  \myrule   
  Feature Maps            & Det.AUCROC     & Loc.AUCROC     & Loc.PRO       & GFLOPs                             \\ \hline
  S23                     & 99.4           & 98.4           & 93.8           & (\textbf{48.0}, 141.8)          \\ 
  S234                    & 98.8           & 98.3           & 92.6           & (59.5, 161.4)                   \\ 
  S123                    & 99.4           & 98.6           & 96.3           & (\textbf{48.0}, 178.8)          \\ \hline
  S123+MP(k=2)            & 99.1           & 98.5           & 95.3           & (\textbf{48.0}, \textbf{44.7})  \\ 
  S123+MP(k=3)            & 99.3           & 98.3           & 94.7           & (\textbf{48.0}, \textbf{44.7})  \\ 
  S123+MP(k=5)            & 99.7           & 98.0           & 93.0           & (\textbf{48.0}, \textbf{44.7})  \\ \hline
  S123+AP(k=2)            & 99.3           & \textbf{98.8}  & 96.8           & (\textbf{48.0}, \textbf{44.7})  \\ 
  \textbf{S123+AP(k=3)}   & 99.7           & \textbf{98.8}  & \textbf{97.1}  & (\textbf{48.0}, \textbf{44.7})  \\
  S123+AP(k=5)            & \textbf{99.8}  & 98.3           & 94.1           & (\textbf{48.0}, \textbf{44.7})  \\ \myrule
  \end{tabular}  
  }}
  \end{table}

\begin{figure}[!htb]
  \centering
  \includegraphics[width=\linewidth]{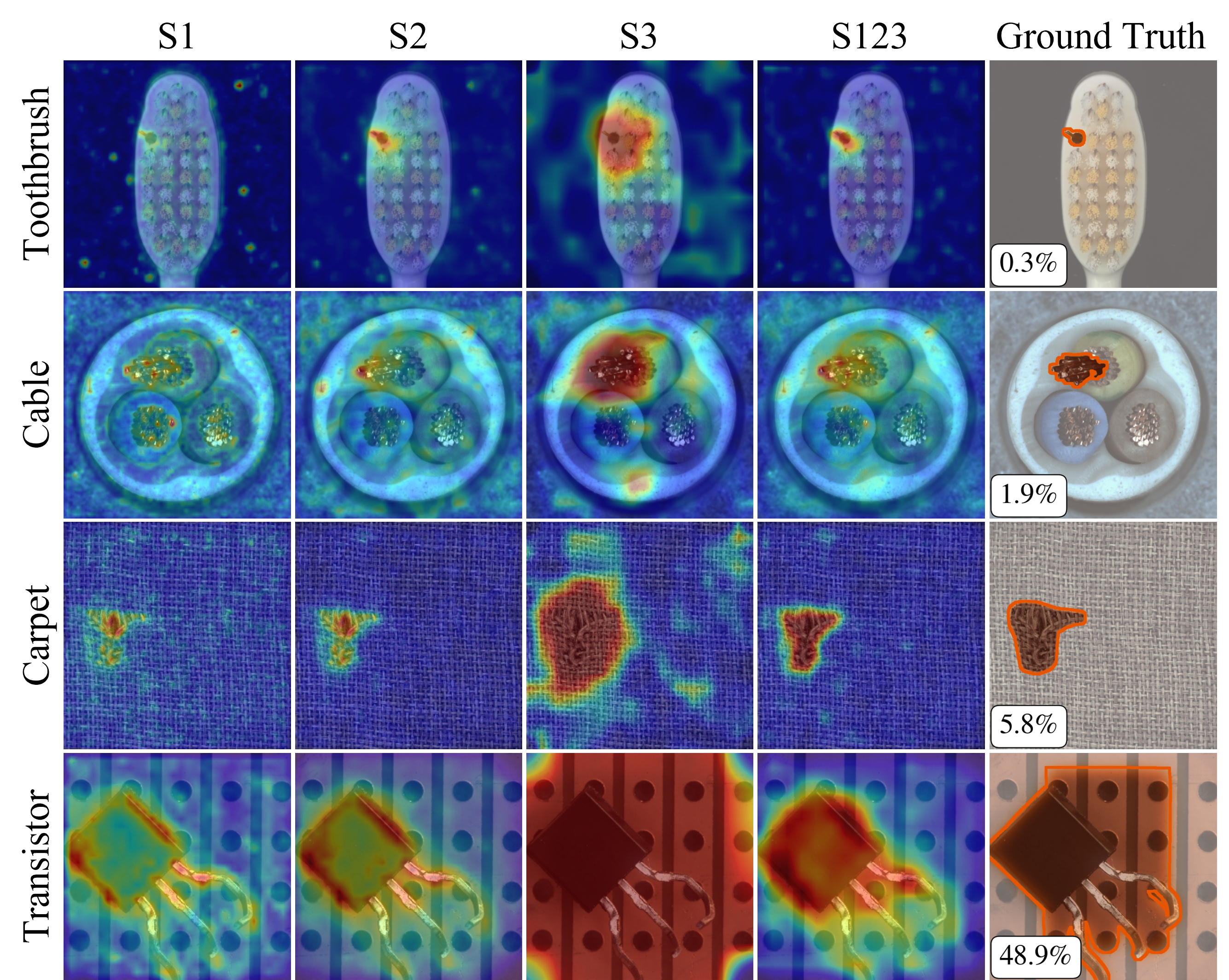}
  \caption{The visualization comparisons of different single-scale and multi-scale outputs for four classes with different defective sizes. The percentages in ground truth images refer to the area proportion of the defects.}
  \label{fig:viz-ss-ms}
\end{figure}

\rounda{
  \textbf{The Ablation on pooling operation}.
  Although only the feature maps of the first 3 stages are extracted to feed into the flow-based model, the GFLOPs of the entire model is still high because of the large-scale feature maps to be estimated. 
	To alleviate the computational burden, pooling layers with stride 2 are used to $2\times$ downsample the extracted feature maps. We also conduct ablation studies about different pooling strategies (max pooling and average pooling) with different kernel sizes (k=2, 3, 5).
	Compared with max pooling, applying the average pooling achieves the same computation reduction yet further boosts the performance, especially for region-wise anomaly localization ($2.4\%$ PRO increment when k=3).
	According to our analysis, the max pooling wildly destroys contextual correlation considering its sample mechanism, which is intolerant to achieving a high region-wise PRO score.
	Shifting the kernel size in the pooling operation also influences the detection performance. Pooling with a small kernel size (k=2) pays attention to details while a large kernel size (k=5) helps to capture global information. 
	Adopting the pooling with kernel size $5$ raises the detection AUROC score to $99.8\%$ but significantly reduces the PRO score to $94.1\%$.
	The detection AUROC score is even up to $99.8\%$ when kernel size is set to $5$ but significantly cut down the PRO localization score. According to the ablation on kernel size as demonstrated in Table \ref{as-feature}, kernel size in the pooling operation is set to $3$ to achieve the detection and localization trade-off.
}

Fig. \ref{fig:viz-ss-ms} straightforwardly displays the localization results of each single-scale and multi-scale on four industrial products, whose defective region sizes vary wildly. When only shage1's feature map is encoded, the flow model locates the minor defects while deriving adverse noises. The noises are alleviated on the feature map of stage2, but the localization results are incomplete. With the feature map of stage3 as the input, the flow model is equipped with the global perception but excessively overrun the localization to non-defect regions. When the feature maps of all scales are encoded, our MSFlow generates precise localization to defective regions regardless of their various sizes. The superior localization results based on feature maps at all scales underscore the significance of multi-scale property for anomaly detection and localization.

\begin{table}[!htb]
  \centering
  \caption{The comparisons of detection and localization performance on MVTec AD with the ResNet-18 as a feature extractor. Some unavailable results of other methods \cite{defard2021padim,gudovskiy2022cflow} are filled by `-'. * refers to our MSFlow based on WRN-50. The best results are highlighted in bold.} \label{tab:as-resnet18}
  \begin{tabular}{l|c|c|c}
  \myrule
  Method                                & Det. AUCROC                  & Loc. AUCROC         & Loc. PRO                             \\ \hline
  RD4AD  \cite{deng2022distillation}    & 97.9                        & 97.1               & 91.2                                 \\ 
  PaDiM  \cite{defard2021padim}        & -                           & 97.1               & 90.8                                 \\ 
  CFlow-AD  \cite{gudovskiy2022cflow}    & 97.1                        & 98.1               & -                                    \\ \hline
  \textbf{MSFlow (Ours)}                    & \textbf{98.9}               & \textbf{98.4}      & \textbf{94.3}                        \\ \hline\hline
  \textbf{MSFlow (Ours)*}               & \textbf{99.7}                & \textbf{98.8}        &  \textbf{97.1} \\ \myrule
  \end{tabular}
  \end{table}

\vspace{3pt}\textbf{Different Feature Extractors.}
We also employ a lightweight backbone ResNet-18 (\rounda{RN-18}) \cite{he2016resnet} as the feature extractor and present a fair comparison with prior methods \cite{deng2022distillation,defard2021padim,gudovskiy2022cflow}, which displayed in Table \ref{tab:as-resnet18}. Based on the feature maps extracted by the RN-18, our MSFlow also remarkably outperforms other methods on both subtasks. Therefore, our MSFlow is flexible and can be treated as a plug-in module with diverse backbones according to application demands. The substantial performance gap between the MSFlow based on RN-18 and WN-50 also reflects the significance of representative features for the flow-based methods.

\subsubsection{Multi-scale Flow Model}
As for the flow model in our MSFlow, the ablation studies are conducted to verify the effectiveness of our asymmetrical parallel flow architecture and fusion flow.

\begin{table}[!htb]
  \centering
  \caption{The ablation study of different numbers of parallel flow blocks. `\#Flow Blocks' denotes the number of flow blocks, where the $i$-th value refers to the flow block number for the feature map of stage $i$. The best results are highlighted in bold.} \label{tab:as-parallel-flow}
  \resizebox{\linewidth}{!}{
  \begin{tabular}{c|c|c|c|c}
  \myrule
  \#Flow Blocks & Det. AUCROC     & Loc. AUCROC     & Loc. PRO       & GFLOPs        \\ \hline
  (5, 5, 5)     & 99.47          & 98.71          & 95.45          & 47.9          \\ 
  (8, 8, 8)     & 99.68          & 98.75          & \textbf{97.34} & 71.1          \\ \hline
  (4, 6, 8)     & 99.69          & \textbf{98.82} & 97.11          & 52.5          \\  
  \textbf{(2, 5, 8)}     & \textbf{99.71} & 98.79          & 97.07          & \textbf{44.7} \\ \myrule
  \end{tabular}}
  \end{table}

\vspace{3pt}\textbf{Asymmetrical Parallel Flows Architecture.} Different from existing flow-based methods \cite{rudolph2022csflow,gudovskiy2022cflow,rudolph2021differnet} stacking as many flow blocks in different branches, our parallel flows adopt an asymmetrical structure to explore the trade-off in performance and efficiency. Table \ref{tab:as-parallel-flow} highlights the advantage of our asymmetrical flow architecture. The obvious performance increments brought by the additional 3 blocks (1st row vs. 2nd row) reveal that the representative capability of flow models can be facilitated with more flow blocks. In our asymmetrical parallel flow architecture, more flow blocks are stacked for the feature map with more channel dimensions. There are two asymmetrical architectures instantiated in the 3rd and 4th row, whose accuracies are similar to the symmetrical architecture of 8 flow blocks (2nd row). According to the 4th row in Table \ref{tab:as-parallel-flow}, it is noteworthy that just 2 flow blocks are enough for the feature map of stage1, and our MSFlow achieves the best trade-off of performance and efficiency in this setting. Our asymmetrical parallel flow architecture avoids the model overfitting on low-dimensional feature maps in symmetric architecture. 

\begin{table}[t]
  \centering
  \rounda{
  \caption{The ablation study of the fusion flow. `+parallels' denotes stacking an additional parallel flow for each scale, and `+fusion' denotes adding one fusion flow after (2, 5, 8) parallel flows. The fusion flow with adaptive feature size $\#a$ is marked as `fusion-s$\#a$'.
  Only the GFLOPs of the flow model are displayed here. The best results are highlighted in bold.} \label{tab:as-fusion-flow}
  \begin{tabular}{l|c|c|c|c}
  \myrule
  flow model            & Det.AUCROC     & Loc.AUCROC     & Loc.PRO       & GFLOPs        \\ \hline
  baseline              & 99.4          & 98.3          & 94.9          & 35.6          \\
  +parallels            & 99.4          & 98.4          & 94.8          & 35.6+7.8      \\ \hline
  +fusion-s8            & 99.3          & 98.5          & 95.2          & 35.6+\tb{7.7} \\
  \textbf{+fusion-s16}  & \textbf{99.7} & \textbf{98.8} & \textbf{97.1} & 35.6+9.1      \\
  +fusion-s32           & 99.1          & 98.3          & 96.0          & 35.6+14.7     \\
  +fusion-s64           & 99.0          & 98.2          & 95.1          & 35.6+36.9     \\\myrule
  \end{tabular}}
  \end{table}

\begin{figure}[h]
  \centering
  \subfloat[The effect of fusion flow on AUCROC scores.]{\includegraphics[width=0.9\linewidth]{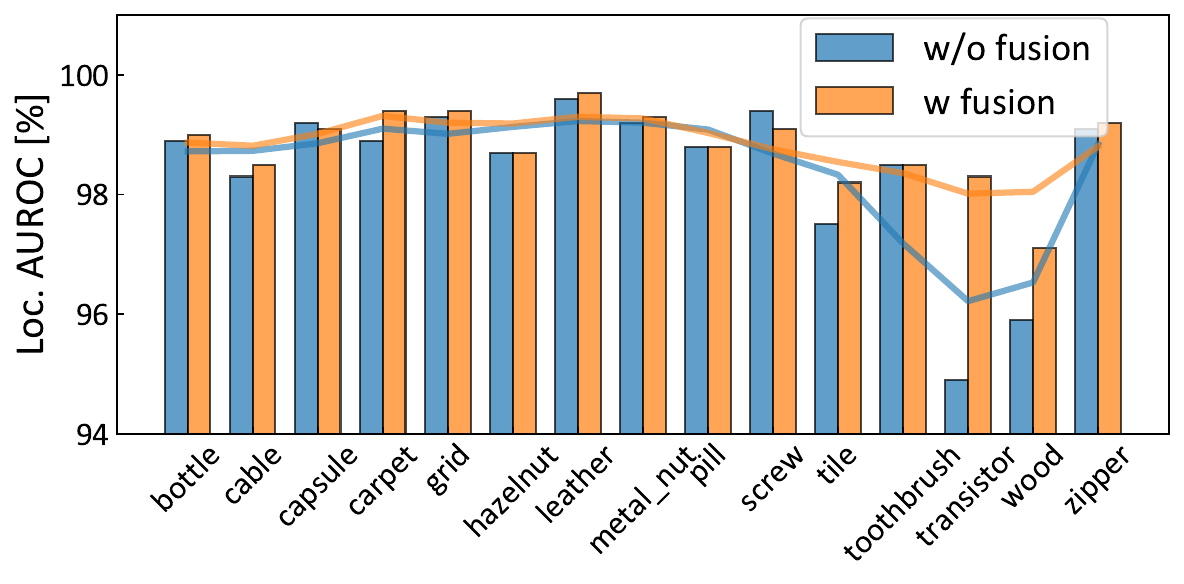}}

  \subfloat[The effect of fusion flow on PRO scores.]{\includegraphics[width=0.9\linewidth]{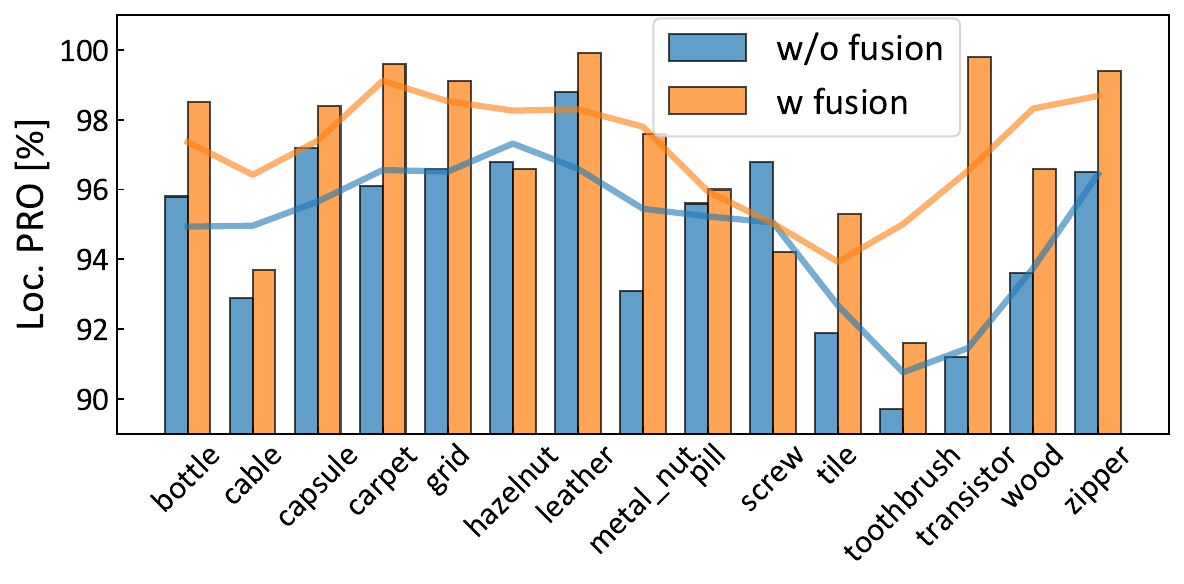}}
  \caption{The effect of fusion flow on AUCROC (a) and PRO (b) scores of all classes in the MVTec AD benchmark. The blue and yellow curves denote the smooth curves of AUCROC scores in all classes with or without the fusion flow.} \label{fig:fusion-all-classes}
\end{figure}

\vspace{3pt}\textbf{The Effect of Fusion Flow.}
Our fusion flow plays an essential role in exchanging spatial perception of different scales and perceptive fields. To verify its effect, the flow model simply dropping the fusion flow is treated as the baseline. Furthermore, because adding a fusion flow to the baseline increases the number of flow blocks, we add one additional flow block for each sequence of parallel flows (3, 6, 9 parallel flow blocks for stage1, stage2, stage3, respectively) for a fair comparison. 
\rounda{
We also conduct ablation studies on the adaptive size in the additional fusion block. The performance and GFLOPs of the flow model are presented in Table \ref{tab:as-fusion-flow}. The additional parallel flow in each branch brings negligible influence on performance. In contrast, our fusion flow facilitates all evaluation metrics, especially the region-wise localization performance PRO score, without introducing heavy computational burdensome. The effect of our fusion flow achieves best when the adaptive size is set to $8$ which is exactly equal to the size of the minimum feature map in $3$ branches.}
With the bridge between different scales built by our fusion flow, the noise of the single scale is suppressed. Therefore, our fusion flow remarkably improves the PRO score that takes connectivity into account by 2.3\%. 
From the effect of fusion flow on AUCROC and PRO scores of all classes in the MVTec AD benchmark as illustrated in Fig. \ref{fig:fusion-all-classes}, our fusion flow is effective and generalized for diverse industrial products with various defect sizes.
\roundb{
  Particularly, when incorporating our fusion flow, there is a discernible decrease in the detection accuracy specifically aimed at identifying \emph{screw}. We conjecture that this decline in performance stems from the consistently minute defects present in \emph{screw} products, as shown in Fig. \ref{fig:defect-examples}. In this scenario, the effectiveness of our fusion flow in covering defect size variation instead becomes a hindrance when attempting to detect these diminutive defects, which are easily overshadowed by the features with the large receptive field.}
  \begin{table}[!htb]
    \centering
    \caption{The detection performance (AUCROC in \%) of CSFlow\cite{rudolph2022csflow} and our FuseFlow. Only the GFLOPs of the flow model are displayed here. The best results are highlighted in bold.} \label{tab:as-fuseflow}
    \rounda{
    \begin{tabular}{l|c|c|c}
    \myrule
    Method   & Det. AUCROC     & \#Params      & GFLOPs       \\ \hline
    CSFlow \cite{rudolph2022csflow}   & \textbf{98.7} & 275.2         & 65.9         \\ 
    \textbf{FuseFlow (Ours)} & 98.6          & \textbf{77.3} & \textbf{4.9} \\ \myrule
    \end{tabular}}
    \end{table}
  
Furthermore, we employ our fusion flow to replace all cross-scale flows in the detection-only CSFlow \cite{rudolph2022csflow} without any other setting changes, and the resulting model is named FuseFlow. From the comparison displayed in Table \ref{tab:as-fuseflow}, our FuseFlow reduces the computational cost of CSFlow by 93\% without sacrificing detection accuracy. This comparison highlights that our fusion flow is much more efficient in information exchange than the cross-scale flow.

\begin{table}[b]
  \centering
  \caption{The ablation study of normalization in flow blocks. w/o Norm refers to dropping the normalization. BN and LN denote batch normalization and layer normalization, respectively. The best results are highlighted in bold.} \label{tab:as-normalization}
  \begin{tabular}{c|c|c|c}
  \myrule
  Normalization & Det. AUCROC    & Loc. AUCROC    & Loc. PRO      \\ \hline
  w/o Norm      & 99.4          & 98.4          & 94.9          \\ 
  w BN          & 99.5          & 98.6          & 95.5          \\ 
  \textbf{w LN}          & \textbf{99.7} & \textbf{98.8} & \textbf{97.1} \\ \myrule
  \end{tabular}
  \end{table}

\vspace{3pt}\textbf{The Effect of Normalizaion in Flow Blocks.}
Compared with the prior flow-based methods \cite{rudolph2022csflow,gudovskiy2022cflow,rudolph2021differnet}, we add a layer normalization (LN) between two $3\times3$ convolutions in st-network of all flow blocks including parallel flows and the fusion flow. The effect of the layer normalization is highlighted in Table \ref{tab:as-normalization}. Removing the normalization decreases the detection AUCROC score by 0.3\% and the localization AUCROC score by 0.4\%. However, the region-wise PRO score drops significantly by 2.2\% without normalization. Although both detection and localization scores are improved when normalization is applied, there is still a performance gap between normalized by BN or LN. This phenomenon can be explained by the different mechanisms of BN and LN, where BN and LN perform statistics in data and model space, respectively.
Because the images are captured on the industrial production lines, the normal training samples of the same industrial class share similar \roundb{pose and appearance}. Therefore, the statistical effect of BN is weakened by the narrow data space, while LN is still effective in the model space.

\subsubsection{Multi-scale likelihood outputs aggregation} This section introduces the ablation studies of different aggregation strategies of the multi-scale likelihood outputs and the image-wise anomaly score calculation. 

\begin{table}[t]
  \centering
  \caption{The ablation study of addition and multiplication aggregation. The best results are highlighted in bold.}  \label{tab:as-aggregation}
  \begin{tabular}{l|c|c|c}
  \myrule
  Aggregation    & Det. AUCROC    & Loc. AUCROC     & Loc. PRO      \\ \hline
  Addition       & 99.4          & \textbf{98.8} & 95.3          \\
  Multiplication & \textbf{99.7} & 98.3          & \textbf{97.1} \\ \myrule
  \end{tabular}
  \end{table}

\begin{figure}[h]
  \centering
  \includegraphics[width=0.9\linewidth]{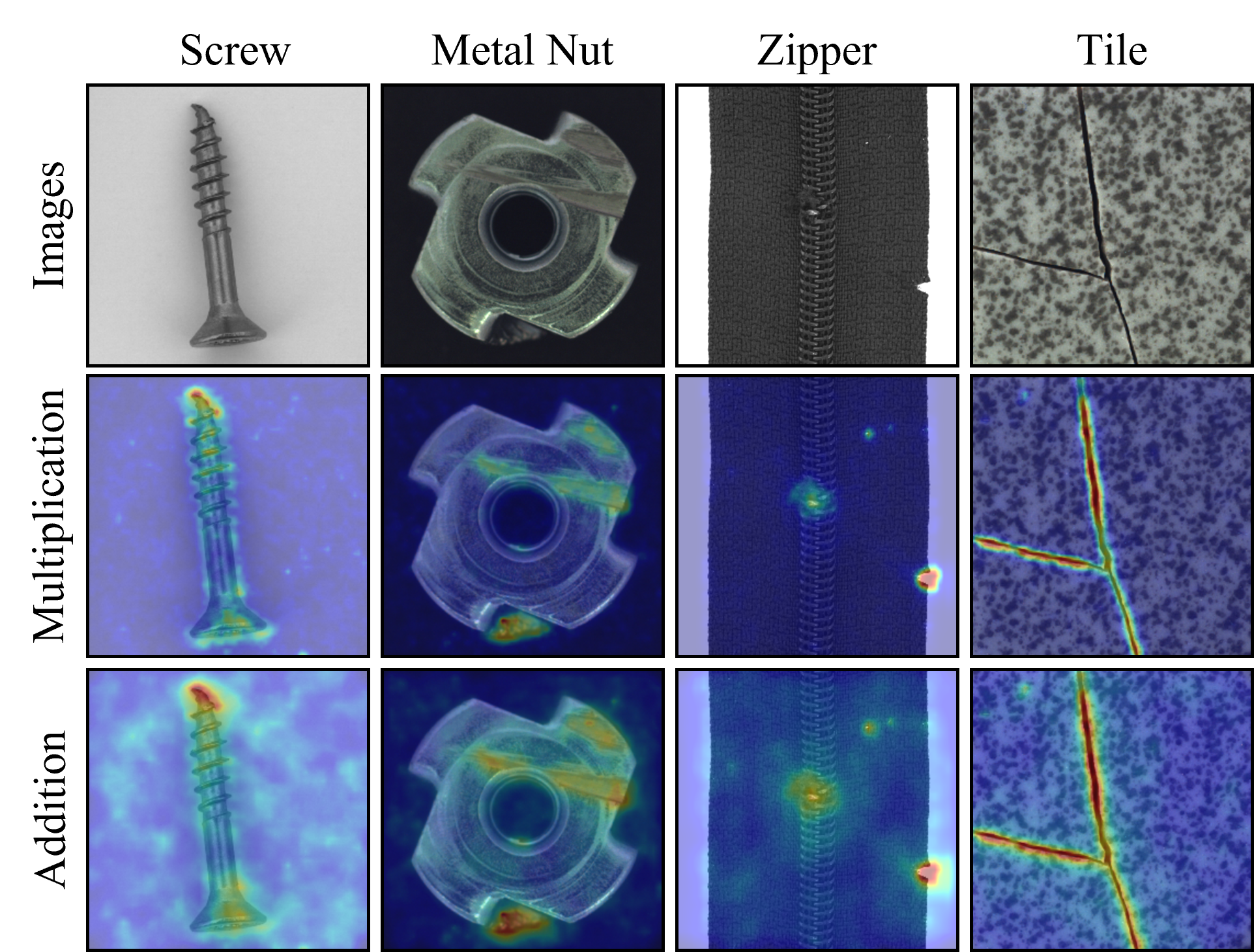}
  \caption{The visualization for anomaly score maps based on multiplication and addition aggregation.}
  \label{fig:add-mul-viz}
\end{figure}

\vspace{3pt}\textbf{Different Aggregation Strategies.}
Considering the inherent discrepancy between anomaly detection and localization, we aggregate multi-scale likelihood maps by common addition for pixel-wise anomaly localization and multiplication for image-wise anomaly localization. Table \ref{tab:as-aggregation} demonstrates the effect of addition and multiplication aggregation on detection and localization performance. As illustrated in Fig. \ref{fig:add-mul-viz}, the multiplication aggregation suppresses noise and only preserves the high likelihoods for regions with high likelihoods across all levels. Therefore, it is suitable for image-wise anomaly score calculation by averaging the largest $K$ anomaly scores and the region-wise PRO score, which takes region connectivity into account. As for the pixel-wise localization AUCROC score, the addition aggregation outperforms the multiplication by retaining more individuality of each scale. 
When using either single aggregation strategy, the proposed MSFlow still outperforms previous methods both in anomaly detection and localization, which demonstrates the superiority of our MSFlow.

\begin{figure}[t]
  \centering
  \includegraphics[width=0.95\linewidth]{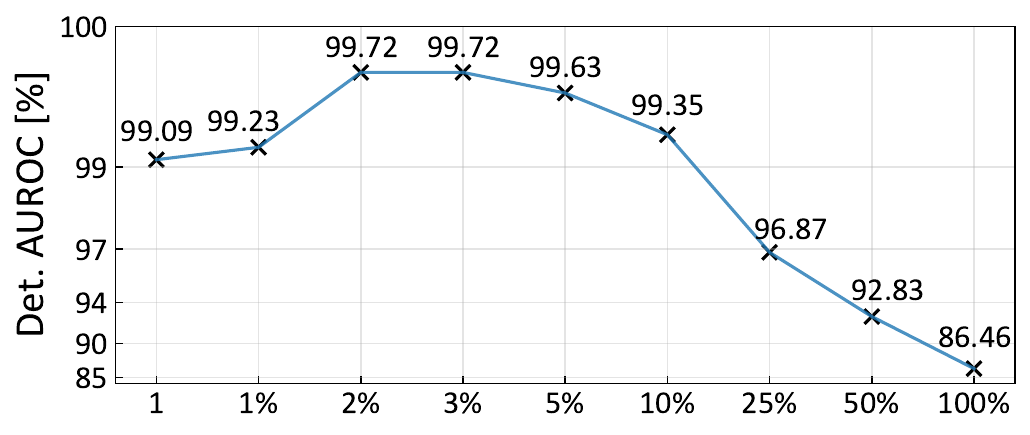}
  \vspace{-10.pt}
  \begin{center}
    $\quad K$
  \end{center}
  \vspace{-5.pt}
  \caption{The effect of different $K$ set in top$K$ on overall detection performance (AUCROC in \%). The mean of top $K$ pixel-wise anomaly scores is viewed as the image-wise anomaly score.}
  \label{fig:as-topk}
\end{figure}

\vspace{3pt}\textbf{The Robustness of the Mean of top\textit{K}.}
To determine the hyperparameter $K$ of top$K$ in image-wise anomaly score calculation, we conduct the ablation study of $K$. Instead of setting a specific value for $K$ directly, we implicitly determine $K$ through the occupancy ratio on the whole anomaly map. The $K$ is respectively set 1 (the maximum), 1\%, 2\%, 3\%, 5\%, 10\%, 25\%, 50\% and 100\% (the mean). Fig. \ref{fig:as-topk} intuitively displays the detection performance tendency of different $K$.
Our MSFlow achieves the best when $K$ is set 2\% or 3\%, and the mean of the top 3\% is treated as the optimum. The detection AUCROC scores remain higher than 99.2\% when $K$ varies from 1\% to 10\%, which reveals the performance is robust to the setting of $K$ in a large range. 

\roundb{
Notably, the performance is dramatically degraded to 86.46\% when $K$ is set to 100\%. The drastic performance degradation seems in conflict with the high performance (98.7\%) achieved by the detection-only method CSFlow \cite{rudolph2022csflow} where the mean is adopted as the image-wise anomaly score. According to our analysis, since our MSFlow achieves high localization performance, only defect regions are assigned with high anomaly scores. Therefore, for the images with minor defects, the mean of the entire anomaly map is misled by the low anomaly scores of most non-defect regions, resulting in such drastic image-wise detection performance degradation.}

\begin{figure}[!htb]
  \vspace{-5pt}
  \centering
  \subfloat[Top 1 (the maximum).]{
    \includegraphics[width=0.4\linewidth]{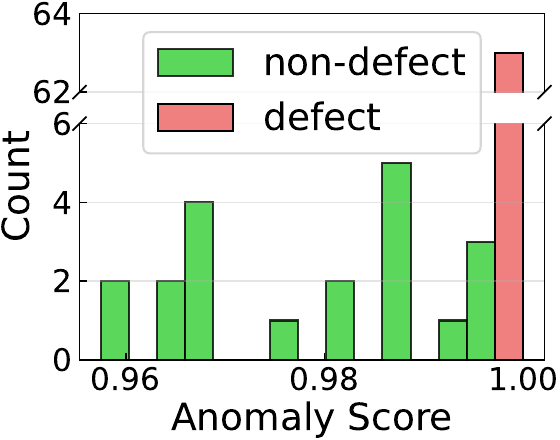}
  }
  \hspace{3mm}
  \subfloat[The mean of top 3\%.]{\includegraphics[width=0.4\linewidth]{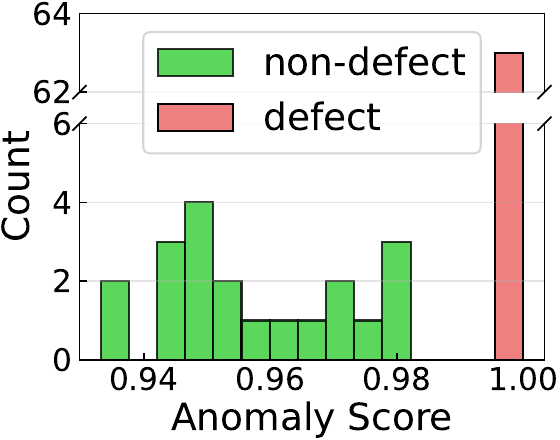}}
  \caption{Image-wise anomaly score distributions based on the maximum and the mean of top 3\% (ours) of `bottle' class in MVTec AD benchmark.} \label{fig:topk-viz}
\end{figure}

To further validate the discriminative capability of the proposed aggregation strategy, we visualize the anomaly score distributions derived from the widely-used maximum (top 1) and our mean of the top 3\%. As as illustrated in Fig. \ref{fig:topk-viz}, although both aggregation approaches achieve 100\% accuracy for for \textit{bottle} class in the MVTec AD benchmark, there is an obvious gap separating defects and non-defects when employing our global anomaly score calculation.

\section{Conclusion} \label{sec:conclusion}
The normalizing flows are appropriate for anomaly detection in an unsupervised fashion. However, the flow-based methods perform inferiorly when facing the unpredictable size variation of anomalies, especially those intending to locate the anomalous regions further. To tackle this issue, this paper exploits the multi-scale potential of flow models and proposes the MSFlow to generalize the anomaly size variation. Our MSFlow achieves state-of-the-art unsupervised anomaly detection with near-perfect 99.7\% detection AUCROC score and 98.8\% localization AUCROC score on the challenging MVTec AD benchmark. The MSFlow also exhibits promising potential for video violation detection. We hope this work will inspire future work about this research interest.

\bibliographystyle{IEEEtran}
\bibliography{ref}

\vspace{-10pt}
\begin{IEEEbiography}
  [{\includegraphics[width=0.95in,height=1.25in]{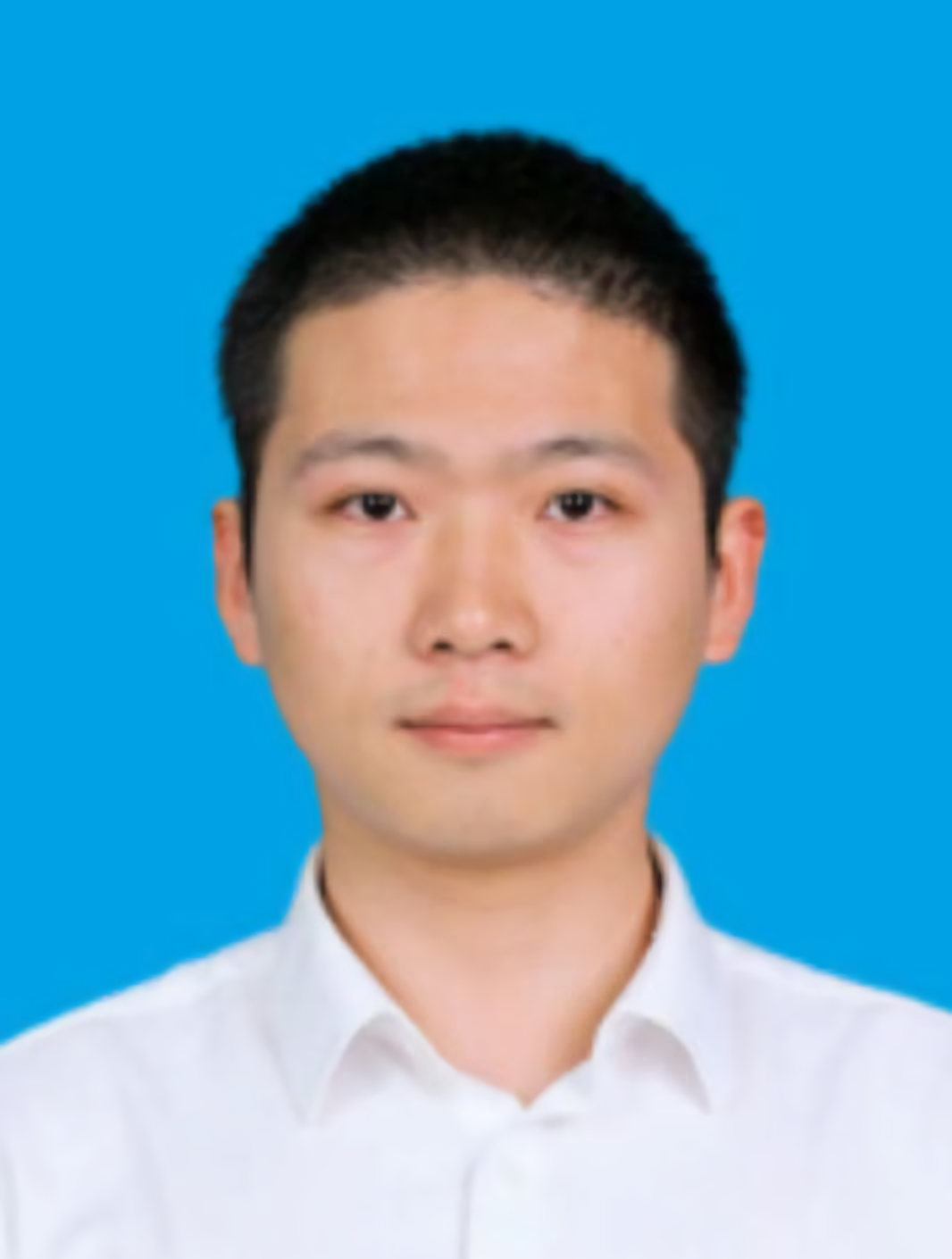}}]
	{Yixuan Zhou}
  received the B.E. degree from the University of Electronic Science and Technology of China, China, in 2019. He is currently pursuing his Ph.D. degree at the Center for Future Media and the School of Computer Science and Engineering, University of Electronic Science and Technology of China, China. His current research interests include unsupervised anomaly detection and computer vision.
\end{IEEEbiography}
\vspace{-10pt}
\begin{IEEEbiography}
  [{\includegraphics[width=1.in,height=1.25in]{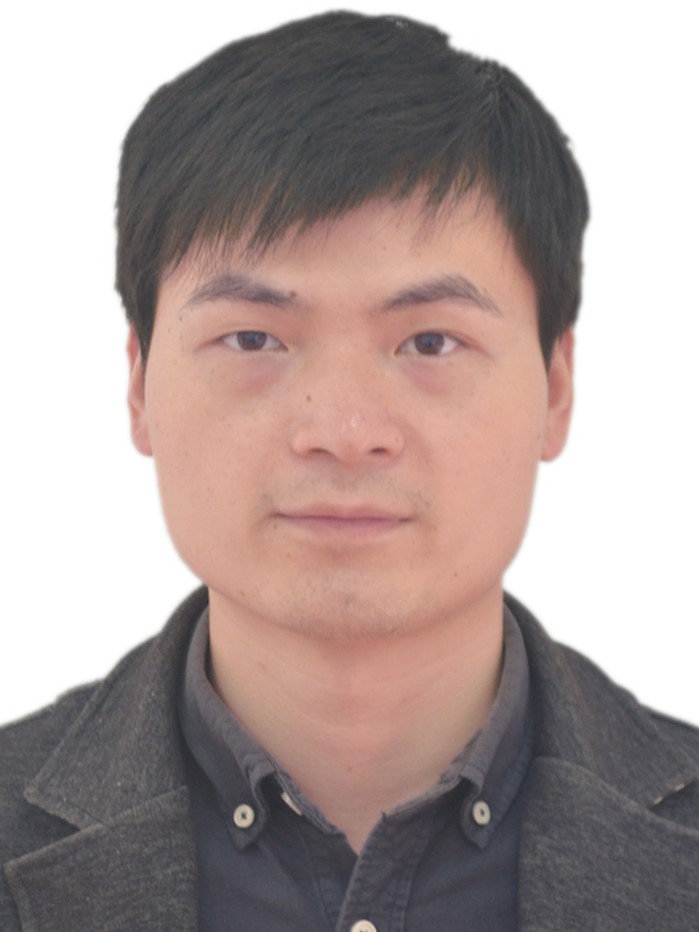}}]
	{Xing Xu}
	received the B.E. and M.E. degrees from Huazhong University of Science and Technology, China, in 2009 and 2012, respectively, and the Ph.D. degree from Kyushu University, Japan, in 2015. He is currently with the School of Computer Science and Engineering, University of Electronic Science and Technology of China, China. He is the recipient of six academic awards, including the IEEE Multimedia Prize Paper 2020, the Best Paper Award from ACM Multimedia 2017, and the World's FIRST 10K Best Paper Award-Platinum Award from IEEE ICME 2017. His current research interests mainly focus on multimedia information retrieval and computer vision.
\end{IEEEbiography}
\vspace{-10pt}
\begin{IEEEbiography}
  [{\includegraphics[width=1.in,height=1.25in]{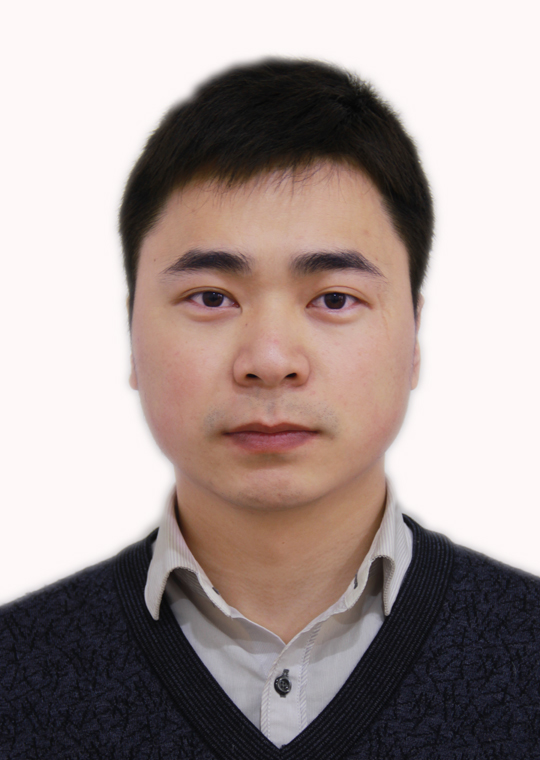}}]
	{Jingkuan Song}
	is a full professor at the University of Electronic Science and Technology of China (UESTC). He joined Columbia University as a Postdoc Research Scientist (2016-2017), and the University of Trento as a Research Fellow (2014-2016). He obtained his Ph.D. degree in 2014 from The University of Queensland (UQ), Australia (advised by Prof. Heng Tao Shen). His research interest includes large-scale multimedia retrieval, image/video segmentation and image/video understanding using hashing, graph learning and deep learning techniques. He was the winner of the Best Paper Award in ICPR (2016, Mexico), the Best Student Paper Award at Australian Database Conference (2017, Australia), and the Best Paper Honorable Mention Award (2017, Japan). He is the Guest Editor of TMM, and WWWJ and is a PC member of CVPR’18, MM'18, IJCAI'18, etc.
\end{IEEEbiography}
\vspace{-10pt}
\begin{IEEEbiography}
  [{\includegraphics[width=1.in,height=1.25in]{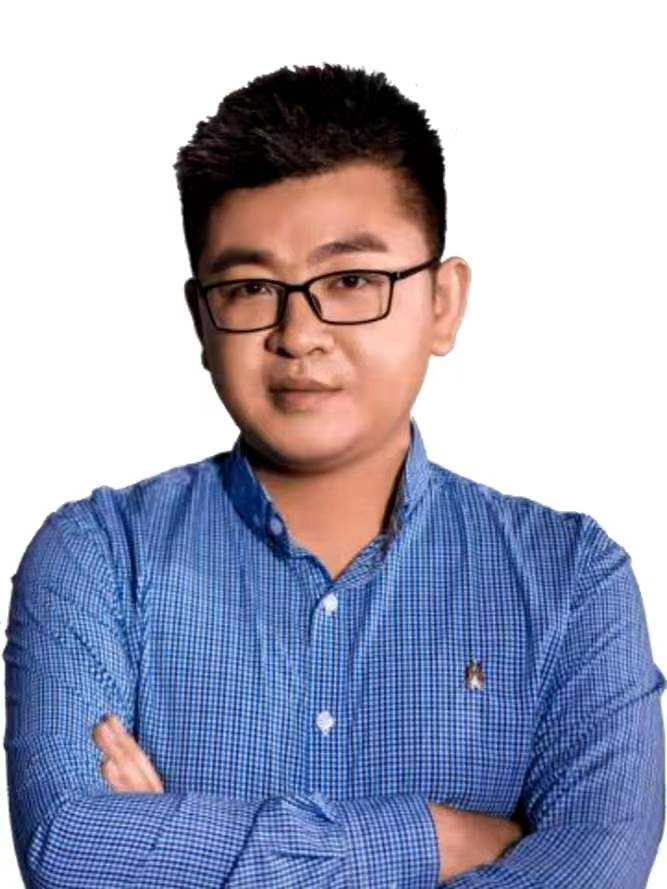}}]
	{Fumin Shen}
	received a bachelor's degree from Shandong University in 2007 and a Ph.D. degree from the Nanjing University of Science and Technology, China, in 2014. He is currently with the School of Computer Science and Engineering, University of Electronic Science and Technology of China. His major research interests include computer vision and machine learning. He was a recipient of the Best Paper Award Honorable Mention from ACM SIGIR 2016 and ACM SIGIR 2017 and the World's FIRST 10K Best Paper Award -- Platinum Award from the IEEE ICME 2017.
\end{IEEEbiography}
\vspace{-10pt}
\begin{IEEEbiography}
  [{\includegraphics[width=1.in,height=1.25in]{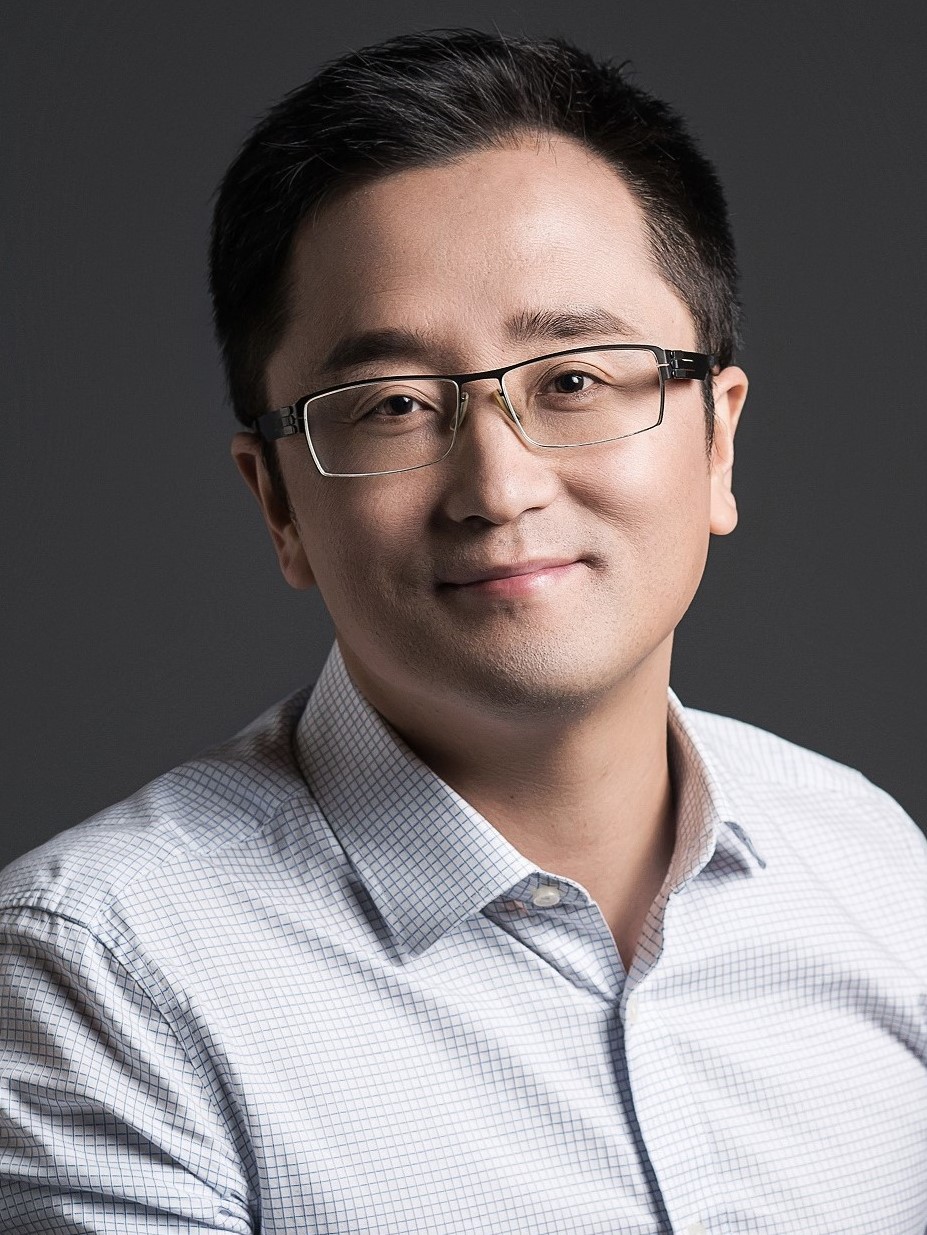}}]
	{Heng Tao Shen} 
	is the Dean of the School of Computer Science and Engineering, and the Executive Dean of the AI Research Institute at the University of Electronic Science and Technology of China (UESTC). He obtained his BSc with 1st class Honours and Ph.D. from the Department of Computer Science, National University of Singapore in 2000 and 2004 respectively. His research interests mainly include Multimedia Search, Computer Vision, Artificial Intelligence, and Big Data Management. He has published 360+ peer-reviewed papers, including 140+ IEEE/ACM Transactions, and received 8 Best Paper Awards from international conferences, including the Best Paper Award from ACM Multimedia 2017 and Best Paper Award - Honourable Mention from ACM SIGIR 2017. He is/was an Associate Editor of ACM Transactions of Data Science, IEEE Transactions on Image Processing, IEEE Transactions on Multimedia, IEEE Transactions on Knowledge and Data Engineering, and Pattern Recognition. He is a Member of Academia Europaea, Fellow of ACM, IEEE and OSA.  
\end{IEEEbiography}

\end{document}